\documentclass[runningheads]{llncs}

\usepackage{eccv}

\usepackage{eccvabbrv}

\usepackage{graphicx}
\usepackage{booktabs}
\usepackage {zebra-goodies}

\usepackage[accsupp]{axessibility}

\PassOptionsToPackage{%
  pagebackref,breaklinks,colorlinks,citecolor=eccvblue
}{hyperref}

\makeatletter
\@ifundefined{endnote}{}{}
\makeatother
\usepackage{skeldoc}

\skelset{hypersetup=colorlinks}

\usepackage{orcidlink}

\usepackage{dsfont}
\usepackage{etoolbox}
\usepackage{color}

\newif\ifshowedits

\newcommand{\addeditor}[3]{%
  \definecolor{#1color}{rgb}{#3}
  \expandafter\newcommand\csname #1\endcsname[1]{%
  \ifshowedits
    {\color{#1color} ##1}%
  \else
    {##1}%
  \fi
  }%
  \expandafter\newcommand\csname #1rmk\endcsname[1]{%
  \ifshowedits
    {\color{#1color} {\bf [#2: ##1]}}
  \fi
  }%
  \expandafter\newcommand\csname #1rpl\endcsname[2]{%
  \ifshowedits
    {\color{#1color} ##1 \sout{##2}}
  \else
    {##1}
  \fi
  }%
}

\newcommand{\createtextvar}[1]{
  \expandafter\newcommand\csname #1\endcsname{%
  {\text{#1}}
}%
}
\newcommand{\mycomment}[1]{}

\newcommand{\IR}{{\mathds{R}}}

\newcommand{\vcomment}[1]{}

\addeditor{ryosuke}{RH}{0.85, 0.5, 0.0}
\addeditor{vincent}{VL}{0.0, 0.5, 0.0}
\addeditor{Ko}{KN}{1,0,1}
\addeditor{kohei}{KY}{0.0, 0.4, 0.6}
\addeditor{antoine}{AG}{0.0, 0.6, 0.4}
\showeditsfalse

\usepackage[utf8]{inputenc}
\usepackage[full]{textcomp}
\usepackage{booktabs} 
\usepackage{multirow} 
\usepackage{pifont}   
\usepackage[table]{xcolor} 
\usepackage{amssymb}  

\usepackage[width=122mm,left=12mm,paperwidth=146mm,height=193mm,top=12mm,paperheight=217mm]{geometry}

\definecolor{best}{RGB}{244, 199, 195}    
\definecolor{second}{RGB}{250, 220, 180}  
\definecolor{third}{RGB}{252, 245, 190}   

\begin{document}


\title{Detailed Geometry and Appearance from Opportunistic Motion} 

\titlerunning{Detailed Geometry and Appearance from Opportunistic Motion}

\author{Ryosuke Hirai\inst{1}\orcidlink{0009-0002-6806-9621} \and
Kohei Yamashita\inst{1}\orcidlink{0000-0002-5086-9906} \and
Antoine Gu{\'e}don\inst{2,3}\orcidlink{0009-0001-3107-4454}\and
Ryo Kawahara\inst{1}\orcidlink{0000-0002-9819-3634}\and
Vincent Lepetit\inst{3}\orcidlink{0000-0001-9985-4433}\and
Ko Nishino\inst{1}\orcidlink{0000-0002-3534-3447}}

\authorrunning{R.~Hirai et al.}

\institute{Graduate School of Informatics, Kyoto University, Japan\\
https://vision.ist.i.kyoto-u.ac.jp/\and
{\'E}cole Polytechnique, France \and
{\fontsize{8}{9}\selectfont LIGM, {\'E}cole Nationale des Ponts et Chauss{\'e}es, IP Paris, Univ Gustave Eiffel, CNRS, France}}

\maketitle

\begin{figure}
\vspace{-12pt}
  \centering
  \includegraphics[width=\linewidth]{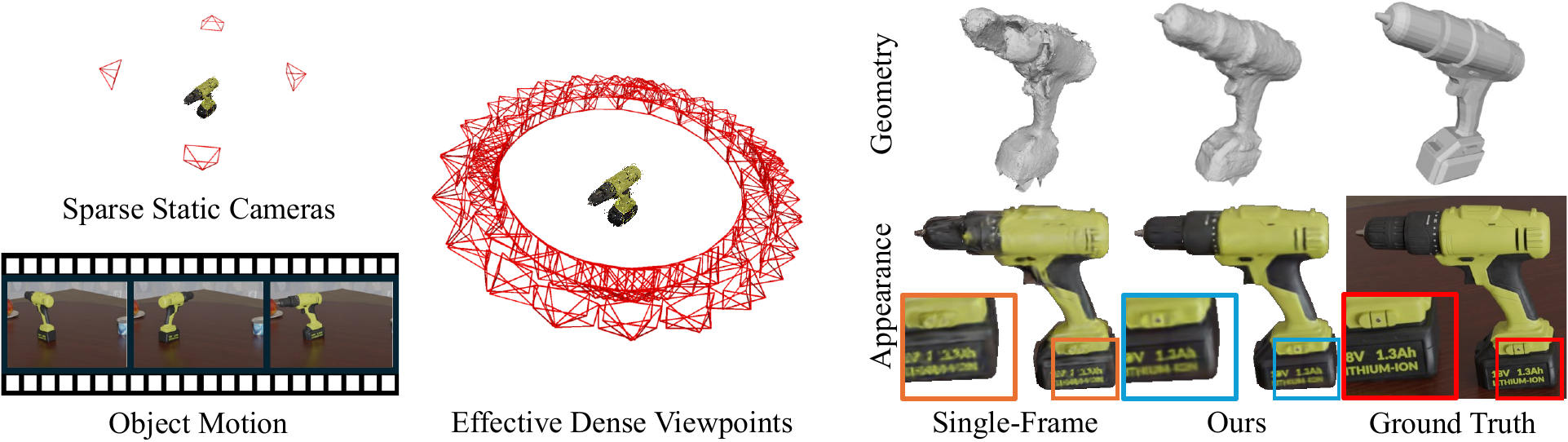} 
  \caption{ 
We introduce a method to recover high-fidelity 3D geometry and appearance from sparse-view static videos. By leveraging ``virtual viewpoints'' induced by opportunistic object motion, our method jointly optimizes object pose and geometry through a motion-aware appearance representation to capture detailed surface structure. 
  }
  \label{fig:opening}
\end{figure}
\vspace{-24pt}

\begin{abstract}
Reconstructing 3D geometry and appearance from a sparse set of fixed cameras is a foundational task with broad applications, yet it remains fundamentally constrained by the limited viewpoints. We show that this bound can be broken by exploiting opportunistic object motion: as a person manipulates an object~(e.g., moving a chair or lifting a mug), the static cameras effectively ``orbit'' the object in its local coordinate frame, providing additional virtual viewpoints. Harnessing this object motion, however, poses two challenges: the tight coupling of object pose and geometry estimation and the complex appearance variations of a moving object under static illumination. We address these by formulating a joint pose and shape optimization using 2D Gaussian splatting with alternating minimization of 6DoF trajectories and primitive parameters, and by introducing a novel appearance model that factorizes diffuse and specular components with reflected directional probing within the spherical harmonics space.
Extensive experiments on synthetic and real-world datasets with extremely sparse viewpoints demonstrate that our method recovers significantly more accurate geometry and appearance than state-of-the-art baselines.

\end{abstract}

\section{Introduction}
\label{sec:intro}

Multiview 3D object reconstruction has achieved a vertical leap in photorealism with the advent of neural representations, most notably 3D Gaussian Splatting~(3DGS)~\cite{kerbl3Dgaussians}. When dense captures from tens of viewpoints are available via a moving camera, these methods recover stunningly accurate geometry and appearance. 
Understanding scenes from a sparse set of fixed viewpoints, however, remains a critical challenge for real-world applications, including home safety monitoring for the elderly or children. The logistical ease of installing static cameras in room corners makes it highly compelling for practical deployment.

Recovering 3D geometry from sparse, static views is inherently ill-posed. The overlap between visual frusta is often insufficient for traditional stereo algorithms, and while monocular estimation can provide coarse structure, it fails to capture surface-level details. 3D Gaussian Splatting also struggles under such extreme sparsity. Recent methods that combine neural depth priors (e.g. MAtCha~\cite{guedon2025matcha}) with neural appearance representations improve accuracy via photometric losses, but as we show in \cref{fig:opening}~(Single-Frame), they still fail to \kohei{reconstruct}
the surface details of individual objects.

If the scene remains entirely static, the information available for reconstruction is strictly bounded by the number of cameras and their spatial baseline, typically yielding crude results with sparse views (\eg, four corner cameras). Fortunately, 
real-world scenes are rarely completely static; the daily activities of a person---picking up a mug, retrieving a book, or moving a chair---unfold continuously. While these movements are often considered a nuisance for 3D reconstruction, they actually provide a wealth of information to recover finer geometric and radiometric details. 

In this paper, we exploit the movement of an object 
to recover higher-resolution geometry and appearance from just a handful of static cameras. 
Two fundamental challenges, however, hamper the use of modern neural representations in this context. First, estimating the relative motion between a target object and a fixed camera is significantly more difficult than standard camera pose estimation. When we have a moving camera, \eg, SLAM or SfM, the entire image including the static background can help determine the camera trajectory. For static cameras with opportunistic object motion, the only cue for estimating the relative camera poses is the object itself. This makes the optimization highly sensitive to the current state of the object geometry.

Second, the standard appearance model in 3DGS is no longer valid. In traditional 3DGS, the radiance of a Gaussian is typically modeled by a set of Spherical Harmonics~(SH) that remain constant over time. This assumes the lighting environment is ``attached'' to the object. When an object moves relative to static light sources, this assumption breaks down. For specular surfaces---common in everyday plastic or metallic objects---the appearance changes drastically as the object rotates relative to the surrounding illumination. Using standard SH coefficients under these conditions results in burned-in lighting and, in turn,corrupt the geometry.

We introduce a novel method to jointly estimate the pose, geometry, and appearance of objects observed by a sparse set of static cameras by leveraging their opportunistic movements. We resolve the coupling between pose and shape through an alternating optimization framework and introduce a motion-aware appearance representation that explicitly models radiance changes during object motion. Our model adopts two physically-grounded assumptions frequently met in real-world scenarios: homogeneous specular reflection and distant lighting. Most everyday objects share a common coating that induces uniform glossiness and are small relative to the distance to environmental light sources. Based on these principles, our appearance model accurately captures complex variations for a wide range of practical applications.

We demonstrate that by correctly modeling these view-dependent and motion-dependent effects, we can leverage object movement to retrieve highly accurate 3D geometry and appearance. We validate our approach on a newly created synthetic dataset and two real-world existing datasets with extremely sparse views, showing that opportunistic motion is a powerful signal for fine-grained reconstruction when handled with a physically-grounded appearance model.

\section{Related Work}

\paragraph{\textbf{3D Scene Reconstruction}} has progressed significantly with neural rendering techniques such as NeRF~\cite{mildenhall2020nerf} and 3D Gaussian Splatting~\cite{kerbl3Dgaussians}.
These methods require dense captures, typically of tens of viewpoints, to optimize scene representation via differentiable rendering.
Subsequent works have further improved rendering quality~\cite{barron2021mipnerf, barron2023zipnerf, Chen2022ECCV, yu_and_fridovichkeil2021plenoxels, mueller2022instant, Yu2024MipSplatting} or focused on extracting explicit surface geometry~\cite{Huang2DGS2024, Yu2024GOF, Dai2024GaussianSurfels, guedon2025milo, wang2021neus, neus2} by leveraging novel representations (\eg, neural SDFs  and 2D Gaussians).
To relax the requirement of dense capture, several methods~\cite{guedon2025matcha, g4splat, long2022sparseneus, ren2023volrecon, wu2023s, raj2025spurfies} tackle sparse-view reconstruction by incorporating learned geometric priors~\cite{long2022sparseneus,ren2023volrecon,wu2023s,guedon2025matcha}.
The accuracy of these methods, however, remains fundamentally bounded by the number of input viewpoints.

\vspace{-8pt}
\paragraph{\textbf{Dynamic 3D Scene Reconstruction}} often employs deformation networks combined with a 3D representations such as neural networks or 3DGS   \cite{pumarola2020d, park2021nerfies, yang2023deformable3dgs} to model non-rigid motion. 
Recent works have specifically focused on extracting surface geometry from these dynamics scenes~\cite{liu2024dynamic, cai2024dynasurfgs, dynamic2dgs}.
These methods, however, typically assume videos with extensive camera motion to provide the necessary multiview information for fine reconstruction. Such requirements limit their applicability in real-world scenarios, such as fixed-camera monitoring. They also often require known camera poses for all moving frames, which is difficult to estimate in dynamic settings. In contrast, our method requires only a sparse set of static, calibrated cameras. 

Another research line focuses on rigidly moving objects to recover both rigid object motion and 3D structure. Although joint pose estimation and reconstruction like Jin \etal~\cite{jin20246dope} and Wen \etal~\cite{bundlesdfwen2023} show promise, they rely on direct depth measurements (RGB-D) which are often unavailable.
In RGB-only settings like ours, current approaches often leverage category-specific priors (\eg, hand-object interaction)~\cite{chen2026forehoi, jiang2025hand}, recover only coarse volumetric density~\cite{wang2024icon, yuan2021star}, or assume highly-textured surfaces for pose tracking~\cite{liu2025h}. Our framework overcomes these limitations by explicitly modeling the specular effects that emerge from the relative motion between lighting, geometry, and cameras. This allows our method to handle texture-sparse objects and achieve high-fidelity 3D reconstructions.

\vspace{-8pt}
\paragraph{\textbf{Appearance Modeling for Specular and Dynamic Objects}}
is indeed challenging for neural object representations. For static objects, recent works have tackled this with specialized radiance functions or inverse rendering for lighting and material parameters~\cite{verbin2022refnerf, ref-neus, ref-gs,jiang2024gaussianshader, gao2024relightable, liu2023nero, kouros2026rgsdr}. 
In dynamic scenes, however, appearance depends on both view and motion. While some methods attempt to handle specularity in dynamic settings~\cite{nerf-ds, fan2025spectromotion}, they typically estimate complex, spatially-varying appearance parameters for every 3D surface point (\eg, each Gaussian primitive). This estimation becomes under-constrained for sparse observations, as in our case. Our approach derives a compact yet expressive appearance model that considers pose-dependent changes in both diffuse and specular reflections. By evaluating spherical harmonics via the reflected view direction, we enable accurate geometry and appearance reconstruction from sparse observations.

\section{Preliminaries: Gaussian Splatting}

Since our method is based on 3D Gaussian Splatting~(3DGS), we briefly review it here while introducing some of the notations we use in the description of our method in the next section.
3DGS models a scene as a collection of $N_g$ anisotropic Gaussian primitives $\{\mathcal{G}_i~|~(i=1,...,N_g)\}$, each defined by its 3D position $\boldsymbol{\mu}_i\in\IR^3$, rotation (quaternion) $\mathbf{q}_i\in\IR^4$, 3D scale $\mathbf{s}_i\in\IR^3_+$, and opacity $\sigma_i\in[0,1]$. To account for view-dependent appearance, each primitive is assigned Spherical Harmonics (SH) coefficients $\theta_i = \{\mathbf{c}_{i,lm}\}_{l=0\dots L,m=-l\dots l}$. For a given viewing direction $\mathbf{v}$, the RGB color $\mathbf{c}_i$ is computed as
\begin{equation}
    \mathbf{c}_i = \mathbf{f}(\mathbf{v}; \theta_i) \, ,
    \> \text{ with } \>
    \mathbf{f}(\mathbf{v}; \theta_i) \equiv \sum_{l=0}^L \sum_{m=-l}^{l} \mathbf{c}_{i,lm} Y_l^m(\mathbf{v}) \, ,
    \label{eq:3dgs_color}
\end{equation}
where $Y_l^m(\mathbf{v})$ denotes an SH basis function.
The Gaussian parameters ($\boldsymbol{\mu}_i$, $\mathbf{q}_i$, $\mathbf{s}_i$, $\sigma_i$, and $\theta_i$) are optimized through a differentiable tile-based rasterizer.

2D Gaussian splatting~\cite{Huang2DGS2024} is a variant for accurate geometry reconstruction.
It represents a 3D scene with oriented planar Gaussian primitives. Each 2D Gaussian is characterized by a surface normal $\mathbf{n}_i$ and a 2D scale, facilitating perspective-consistent depth evaluation and accurate surface extraction. 
These surface normals allow for direct geometric regularization---such as depth-normal consistency losses---and, as we show later, provide the necessary geometric context to model complex view- and pose-dependent appearance for moving objects.

\begin{figure*}[t]
  \centering
  \subfloat[Inputs]{
    \includegraphics[height=0.24\textwidth]{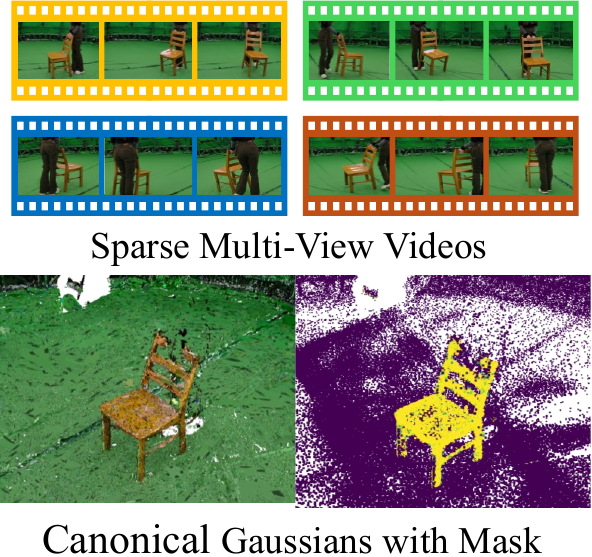}
  }
  \hspace{0.1em}\rule{0.6pt}{0.24\textwidth}\hspace{-0.1em}
  \subfloat[Object pose estimation]{
    \includegraphics[height=0.24\textwidth]{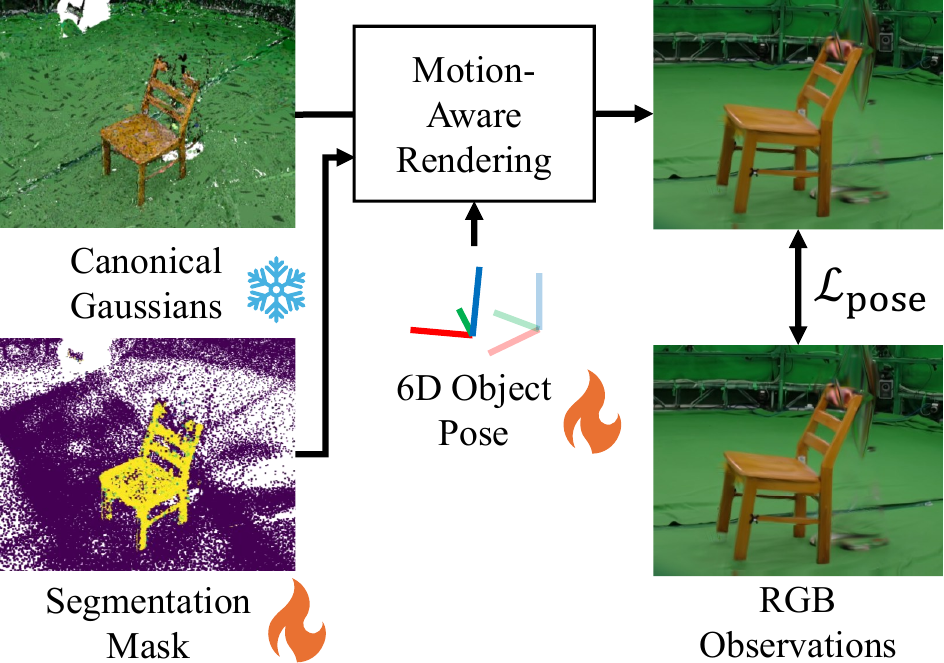}
  }
  \hspace{0.1em}\rule{0.6pt}{0.24\textwidth}\hspace{-0.1em}
  \subfloat[Gaussian Refinement]{
    \includegraphics[height=0.24\textwidth]{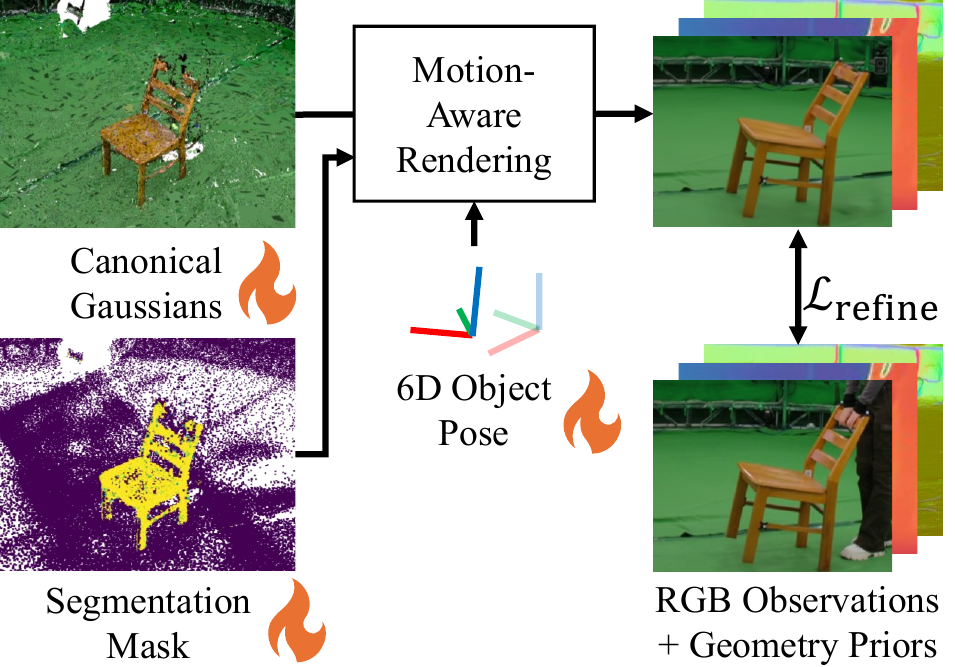}
  }
  \caption{
Method Overview. Given sparse multi-view videos and an initial set of canonical 2D Gaussians (a), our framework recovers per-frame object poses while iteratively refining the Gaussian geometry via differentiable rendering. To ensure robust convergence under sparse supervision, we employ an alternating optimization that switches between 6-DoF object pose estimation (b) and canonical Gaussian refinement (c) using the aggregated temporal information from all processed frames.
  }
  \label{fig:overview}
\end{figure*}

\section{Method}
\label{sec:method}

Given sparse-view videos captured by $V$ static cameras (\eg, $V=4$ corner cameras), our goal is to reconstruct the geometry and appearance of a rigidly moving object with high-fidelity. 
The system inputs consist of RGB frames $\{\mathbf{I}_v^t~|~v=1,...,V,t=1,...,T\}$, camera extrinsics $\mathbf{T}_v~(v=1,...,V)$, 
and intrinsics $\mathbf{K}_v~(v=1,...,V)$, where $v$ and $t$ denote camera and temporal indices, respectively. For appearance modeling, we assume objects are sufficiently small relative to their environment (\eg, a room) to justify a distant-lighting approximation. We also empirically assume uniform gloss, as the specular properties of many everyday objects are dominated by a consistent surface coating. Relaxing these assumptions represents a meaningful direction for future work.

We initialize the whole 3D scene at $t=1$ as a set of ``canonical'' 2D Gaussians $\{\mathcal{G}_i~|~(i=1,...,N_g)\}$ using $\{\mathbf{I}_v^1\}$ and learned priors (\cref{sec:initialization}). 
To isolate the moving object from the static environment, we estimate a soft segmentation mask, $m_i \in [0,1]$, for each Gaussian, where $m_i$ = 0 represents the background and $m_i=1$ represents the rigidly moving foreground. Final surface geometry is extracted from these canonical Gaussians via post-processing meshing (\cref{sec:meshing}).

We exploit the dense-view observations induced by object motion by jointly optimizing the object's trajectory and the canonical Gaussians. For each timestep $t$, the object pose is parameterized by a quaternion $\mathbf{q}_\mathrm{obj}^t$ and a 3D translation $\mathbf{t}_\mathrm{obj}^t$. Using differentiable rendering, we iteratively refine these pose parameters and the Gaussian attributes by minimizing the photometric discrepancy between rendered and observed video frames. 

Joint optimization of pose and geometry is, however, significantly more sensitive to initialization than standard SLAM and SfM for static scenes. Under extreme view-sparsity, relying on limited photometric cues can lead to unstable convergence. To ensure robustness, we propose an alternating optimization framework~(\cref{fig:overview}). Furthermore, to better leverage radiometric cues, we introduce a compact yet expressive appearance model that accounts for view- and pose-dependent appearance changes throughout the sequence.

\subsection{Single-Frame 3D Reconstruction with Learned Priors} \label{sec:initialization}

We bootstrap our alternating estimation by initializing the canonical Gaussians from the first frames of the $V$ views using MAtCha Gaussians~\cite{guedon2025matcha}, a method that recovers scene geometry and appearance of static scenes from sparse multi-view images. 
MAtCha first estimates depth and normal maps for each view by leveraging monocular depth estimation~\cite{depth_anything_v2} and learning-based structure-from-motion~\cite{duisterhof2025mastrsfm}. 
It then optimizes a set of 2D Gaussians supervised by input images and regularized by these geometric priors.

We provide the multi-view images at the first timestep, $\{\mathbf{I}_v^1|v=1, \dots, V\}$, along with the camera parameters to the MAtCha pipeline to obtain our initial canonical Gaussians. We also run MAtCha at each subsequent timestep to generate per-frame and per-view geometric priors, which serve as regularization during our alternating optimization process. 

To isolate the target object, we initialize the mask values $m_i$ for each Gaussian using 2D segmentation masks from the initial timestep, generated via SAM~2~\cite{ravi2024sam2}.  We optimize these mask values by rendering a mask image---assigning $m_i$ as the color for each Gaussian---and minimizing the photometric loss against the pseudo ground-truth masks.
Since we have ground truth segmentation masks for synthetic data, we use them instead of SAM-2 segmentation masks.

\subsection{Motion-Aware Joint Optimization via Soft-Masked Transform}

Using multi-view observations from subsequent timesteps, we alternate between estimating per-frame object poses and refining the canonical Gaussian parameters. In both stages, we transform the canonical Gaussians associated with the foreground (those with high mask values $m_i$) according to the current pose estimate, render the scene, and back-propagate photometric gradients.

A significant challenge is that initial mask values may be noisy, potentially degrading estimation accuracy. We address this by refining the per-Gaussian mask values $m_i$ concurrently with object motion. To facilitate this, we derive a soft-masked rigid transformation that allows gradients to flow directly to the mask values. Given the object pose $(\mathbf{q}_\mathrm{obj}^t, \mathbf{t}_\mathrm{obj}^t)$ and mask values $m_i \in [0,1]$, we compute the interpolated motion for each Gaussian:
\begin{equation}
    \mathbf{q}_{\mathrm{obj},i}^t = \mathrm{Normalize}[m_i \cdot \mathbf{q}_\mathrm{obj}^t + (1-m_i)\cdot \mathbf{q}_\mathrm{I}]\,,\text{ and}
\end{equation}
\begin{equation}
    \mathbf{t}_{\mathrm{obj},i}^t = m_i\cdot \mathbf{t}_\mathrm{obj}^t\,,
\end{equation}
where $\mathbf{q}_\mathrm{I}$ is the identity rotation $[1,0,0,0]$.

The canonical Gaussians $\{\mathcal{G}_i\}$ are then transformed to the current state $\{\mathcal{G}_i^t\}$ by updating their spatial parameters. Specifically, the transformed means $\boldsymbol{\mu}_i^t$ and orientations $\mathbf{q}_i^t$ are computed as:
\begin{equation}\label{eq:mu_t}
    \boldsymbol{\mu}_i^t = \mathbf{R}_{\mathrm{obj},i}^t\left(\boldsymbol{\mu}_i - \mathbf{p}\right) + \mathbf{t}_{\mathrm{obj},i}^t + \mathbf{p}\,,\text{ and}
\end{equation}
\begin{equation}\label{eq:q_t}
    \mathbf{q}_i^t = \mathbf{q}_{\mathrm{obj},i}^t \cdot \mathbf{q}_i\,,
\end{equation}
where $\mathbf{R}_{\mathrm{obj},i}^t$ is a rotation matrix for $\mathbf{q}_{\mathrm{obj},i}^t$ and $\mathbf{p}$ is the object centroid in the canonical space computed from the initial canonical Gaussians and the initial mask values.
Other Gaussian attributes remain fixed to their canonical values. For this phase, we use the standard appearance model (\cref{eq:3dgs_color}), but evaluate the viewing direction in the object's local coordinate system. This effectively assumes a ``body-attached'' lighting environment, which we further refine in \cref{sec:modeling_specular_effect}.

We supervise the optimization using a composite loss function. Following 3DGS~\cite{kerbl3Dgaussians} and 2DGS~\cite{Huang2DGS2024}, we employ a rendering loss $\mathcal{L}_\mathrm{RGB}$ (comprising $L_1$ and D-SSIM) and a depth-normal consistency loss $\mathcal{L}_n$. To ensure stability under sparse views, we integrate the geometric priors from MAtCha~\cite{guedon2025matcha}: a depth prior loss $\mathcal{L}_\mathrm{pdepth}$ and a normal prior loss $\mathcal{L}_\mathrm{pnormal}$ based on foundation model estimates~\cite{depth_anything_v2,duisterhof2025mastrsfm}. Finally, to enforce a clear separation between the moving object and the static background, we apply an entropy-based binary regularization on the mask values:
\begin{equation}\label{eq:entropy}
    \mathcal{L}_\mathrm{entropy} = \sum_i\{-m_i\mathrm{log}m_i - (1-m_i)\mathrm{log}(1-m_i)\}\,,
\end{equation}
for physically meaningful estimation.
The overall loss function for the pose estimation phase $\mathcal{L}_\mathrm{pose}$ and Gaussian refinement phase $\mathcal{L}_\mathrm{refine}$ are defined as follows:
\begin{equation}
    \mathcal{L}_\mathrm{pose} = \mathcal{L}_\mathrm{RGB} + \lambda_\mathrm{entropy}\mathcal{L}_\mathrm{entropy} \,, \text{ and}
\end{equation}
\begin{equation}
    \mathcal{L}_\mathrm{refine} = \mathcal{L}_\mathrm{RGB} + \lambda_\mathrm{n}\mathcal{L}_\mathrm{n} + \lambda_\mathrm{pnormal}\mathcal{L}_\mathrm{pnormal} + \lambda_\mathrm{pdepth}\mathcal{L}_\mathrm{pdepth} + \lambda_\mathrm{entropy}\mathcal{L}_\mathrm{entropy} \,.
\end{equation}

\begin{figure*}[t]
  \centering  
  \subfloat[Effective Light for Specular Reflection]{
    \includegraphics[height=0.15\textwidth]{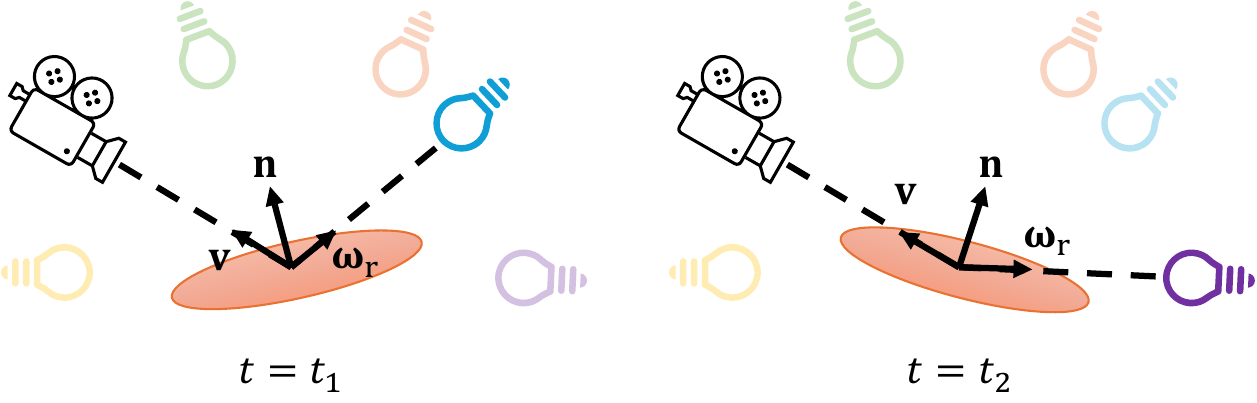}
    \label{fig:specular_reflection}
  }
  \hfill
  \subfloat[Effective Hemisphere for Diffuse Shading]{
    \includegraphics[height=0.15\textwidth]{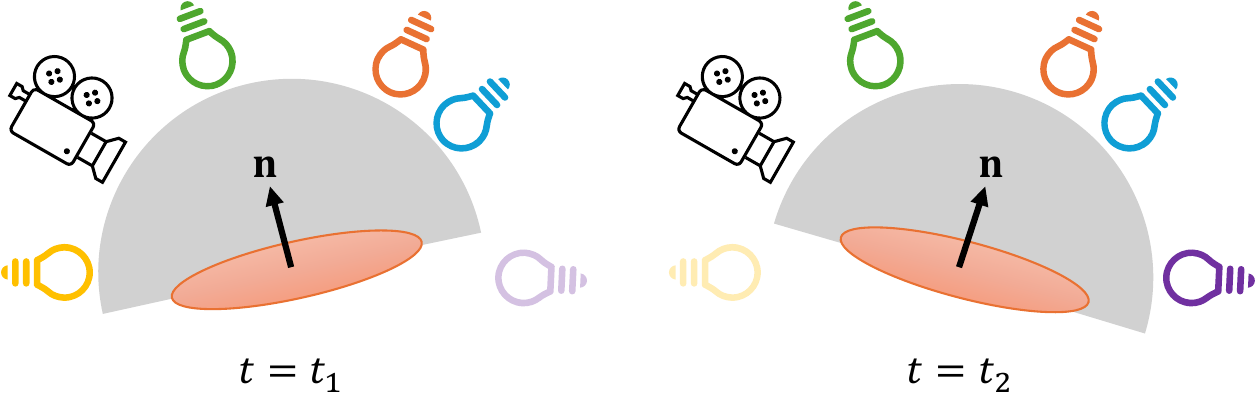}
    \label{fig:diffuse_reflection}
  }
  \\
  \subfloat[Definition of Inputs]{
    \includegraphics[height=0.18\textwidth]{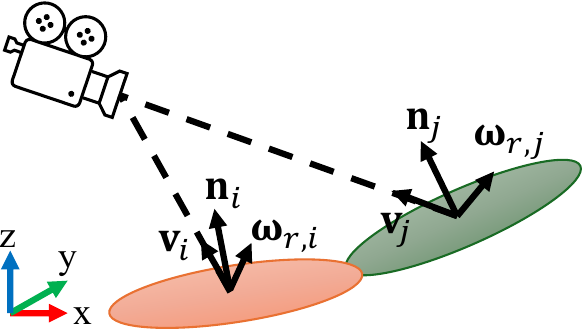}
  }
  \subfloat[3DGS]{
    \includegraphics[height=0.18\textwidth]{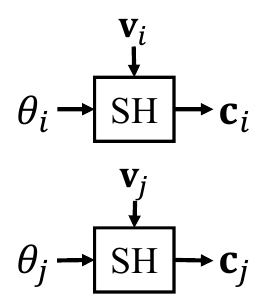}
  }
  \subfloat[Specular Model]{
    \includegraphics[height=0.18\textwidth]{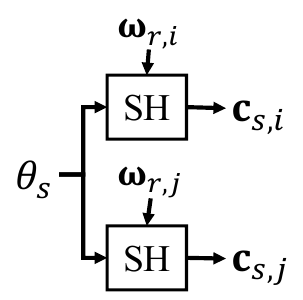}
  }
  \subfloat[Diffuse Model]{
    \includegraphics[height=0.18\textwidth]{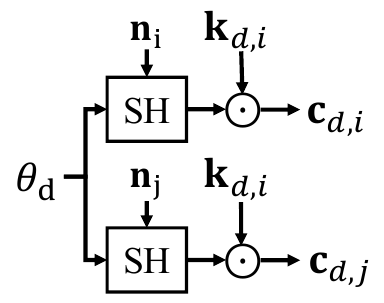}
  }
  \caption{
\textbf{Motion-Aware Appearance Modeling.} (a) For moving objects, specular reflection is a function of incident radiance from the reflected viewing direction $\boldsymbol{\omega}_r$, which evolves with object pose. (b) Similarly, diffuse reflection depends on the time-varying surface normal $\mathbf{n}$ relative to the static environment. (c) Our model factorizes appearance into specular (e) and diffuse (f) components by evaluating the surface normal and reflected viewing directions in the world coordinate system. (d) Unlike standard 3DGS, which optimizes independent Spherical Harmonics (SH) for each primitive, our approach employs shared SH coefficients $\theta_d$ and $\theta_s$ across all foreground Gaussians to robustly capture the global illumination field.
  }
  \label{fig:appearance_modeling}
\end{figure*}

\subsection{Motion-Aware Appearance Modeling via Radiance Probing}
\label{sec:modeling_specular_effect}

Following the alternating optimization phase, we perform a joint refinement of geometry and appearance using the full temporal sequence. To achieve high-fidelity reconstruction from sparse views, we derive a compact, physically-inspired appearance model tailored for moving objects.

As illustrated in \cref{fig:specular_reflection}, specular radiance on smooth surfaces primarily depends on the surface reflectance $\mathbf{k}_s$ and the incident illumination from the reflected viewing direction:
\begin{equation}
    \boldsymbol{\omega}_r = -\mathbf{v} + 2 \left(\mathbf{v} \cdot \mathbf{n} \right) \mathbf{n} \,,
\end{equation}
where $\mathbf{v}$ and $\mathbf{n}$ denote the viewing direction and surface normal in the world coordinate system, respectively. 
Let $\odot$ denote the Hadamard product.
We approximate the specular component for each Gaussian ($i$) as
\begin{equation}
    \mathbf{c}_{s,i} = \mathbf{k}_{s,i} \odot \mathbf{f}({\boldsymbol{\omega}}_{r,i}; \theta_s) \,,
\end{equation}
where $\mathbf{k}_{s,i}$ is the specular reflectance and ${\boldsymbol{\omega}}_{r,i}$ is the view direction reflected by the surface normal $\mathbf{n}_i$. In effect, this models a Phong-like~\cite{phong75} specular component. $\mathbf{f}$ is a weighted sum of SH basis functions. The learnable SH coefficients $\theta_s$ approximate the incident radiance from the surrounding environment probed by the reflected view direction. Assuming the illumination is distant relative to the object size, the incident radiance remains invariant to surface translation; thus, we employ a single set of SH coefficients $\theta_s$ shared across all foreground Gaussians. In practice, we set $\mathbf{k}_{s,i} = 1$, \ie, assume objects with uniform gloss.

We further account for diffuse reflection, which also exhibits time-dependency as the object rotates through a static lighting environment. The radiance for a Lambertian surface is defined by 
\begin{equation}
    \mathbf{c}_d = \mathbf{k}_d \odot \int \boldsymbol{L}_i(\boldsymbol{\omega}_i) \max\left(\boldsymbol{\omega}_i \cdot \mathbf{n}, 0\right) \mathrm{d}\boldsymbol{\omega}_i \,,
\end{equation}
where $\mathbf{k}_d$ is diffuse albedo and $\boldsymbol{L}_i$ denotes the incident radiance as a function of incident direction $\boldsymbol{\omega_i}$. As depicted in \cref{fig:diffuse_reflection}, this integral is a function of the time-varying surface normal $\mathbf{n}$. We approximate the diffuse component as  
\begin{equation}
    \mathbf{c}_{d,i} = \mathbf{k}_{d,i} \odot \mathbf{f}({\mathbf{n}}_i; \theta_d) \,,
\end{equation}
where $\theta_d$ represents the learnable SH coefficients for the diffuse irradiance. 

Our final appearance model computes the total color as sum of the specular and diffuse components:
\begin{equation}
    \mathbf{c}_i = \mathbf{f}({\boldsymbol{\omega}}_r; \theta_s) + \mathbf{k}_{d,i} \odot \mathbf{f}(\hat{\mathbf{n}}_i; \theta_d) \,.
    \label{eq:appearance_model}
\end{equation}

\Cref{fig:appearance_modeling} (c)-(f) summarizes the differences between the standard 3DGS appearance model and our motion-aware factorized appearance model. 
By sharing SH coefficients across all foreground primitives and only optimizing per-Gaussian albedo, our model remains highly expressive yet robust to sparse supervision. Unlike standard 3DGS, which optimizes independent SH coefficients for every primitive, our model explicitly leverages the geometric relationship between surface normals and world-space illumination.

In this final refinement stage, we omit the monocular depth and normal prior losses, as the accumulated temporal observations provide sufficient geometric constraints. The final optimization objective is:
\begin{equation}
    \mathcal{L}_\mathrm{g} = \mathcal{L}_\mathrm{RGB} + \lambda_\mathrm{normal}\mathcal{L}_\mathrm{n}  + \lambda_\mathrm{entropy}\mathcal{L}_\mathrm{entropy} \,.
\end{equation}

\subsection{Meshing with Viewpoints from All Frames} 
\label{sec:meshing}

Once the joint optimization is complete, we extract a surface mesh of the foreground object. We adapt the strategy of MILo~\cite{guedon2025milo} to our dynamic setting, allowing the meshing process to leverage observations of the motion at all timesteps rather than relying solely on the canonical frame.

We first isolate the foreground Gaussians by thresholding the optimized mask values $m_i$. Each selected Gaussian $\mathcal{G}_i$ spawns a fixed set of pivot points anchored within its local coordinate frame.
Because pivots are anchored to their parent Gaussian, they follow the object through time: at timestep $t$, each pivot is transformed by the same masked rigid motion as its Gaussian (\cref{eq:mu_t,eq:q_t}).
A Delaunay triangulation of the pivots yields a volumetric tetrahedral mesh which evolves over time as the pivots move with the object.

We assign a learnable signed distance value (SDF) to each pivot and optimize these values over 1000 iterations.
At each iteration, we first select a timestep and extract a triangle mesh from the current SDF values and Delaunay triangulation via differentiable marching tetrahedra~\cite{e74-d_1_214}. Then, we select a viewpoint, render depth and normal maps from the extracted mesh, and compare them to the corresponding depth and normal maps rendered from the Gaussians.
Gradients are back-propagated from this rendering loss to the SDF values, enforcing the surface of the dynamic object to match the geometry of the Gaussians.

Crucially, this supervision aggregates over all views $v$ and all timesteps $t$, so that the SDF values benefit from the same effectively dense observations that drove the Gaussian refinement.
After convergence, the final mesh is extracted by applying marching tetrahedra to the first frame pivot positions.

\section{Experimental Results}
\label{sec:experiments}

\begin{table}[t]
\caption{
\textbf{Surface normal accuracy for synthetic data.} We evaluate reconstruction quality on synthetic data using the mean, median, and 80th percentile errors against ground-truth surface normals. The results demonstrate that our proposed components significantly enhance the recovery of fine surface details. While simpler surfaces, such as the drain cleaner, can be handled by baseline methods, they do not fully reflect our method's capability to reconstruct intricate geometric details.
}
\label{tab:synth_normal_quantitative}
\centering
\small
\resizebox{\textwidth}{!}{%
\setlength{\tabcolsep}{2.pt} 

\begin{tabular}{l *{18}{r}}
\toprule
\multirow{2}{*}{Method}
& \multicolumn{3}{c}{drain cleaner}
& \multicolumn{3}{c}{ottoman}
& \multicolumn{3}{c}{bunny}
& \multicolumn{3}{c}{garden gnome}
& \multicolumn{3}{c}{drill}
& \multicolumn{3}{c}{Average} \\
\cmidrule(lr){2-4}\cmidrule(lr){5-7}\cmidrule(lr){8-10}\cmidrule(lr){11-13}\cmidrule(lr){14-16}\cmidrule(lr){17-19}
& Mean $\downarrow$ & Med. $\downarrow$ & P80 $\downarrow$
& Mean $\downarrow$ & Med. $\downarrow$ & P80 $\downarrow$
& Mean $\downarrow$ & Med. $\downarrow$ & P80 $\downarrow$
& Mean $\downarrow$ & Med. $\downarrow$ & P80 $\downarrow$
& Mean $\downarrow$ & Med. $\downarrow$ & P80 $\downarrow$
& Mean $\downarrow$ & Med. $\downarrow$ & P80 $\downarrow$ \\
\midrule
DG-Mesh~\cite{liu2024dynamic} & 28.3 & 24.5 & 41.7 & 26.9 & 22.7 & 39.2 & 33.8 & 30.5 & 49.3 & 26.0 & 22.3 & 38.6 & 26.9 & 23.6 & 39.3 & 28.4 & 24.7 & 41.6 \\
\midrule
Single-Frame
& 11.8 & \cellcolor{third}7.2 & 16.3
& 18.1 & \cellcolor{third}13.0 & \cellcolor{third}20.0
& 18.4 & 14.4 & 25.9
& 22.9 & 16.5 & 35.2
& 21.3 & 15.7 & 32.5
& 18.5 & 13.4 & 26.0 \\
w/o Motion-Aware Appearance
& \cellcolor{best}8.4 & \cellcolor{best}4.8 & \cellcolor{second}12.0
& \cellcolor{third}17.9 & \cellcolor{third}13.0 & \cellcolor{third}20.0
& 15.3 & \cellcolor{third}12.4 & 21.6
& \cellcolor{second}20.7 & \cellcolor{third}15.9 & \cellcolor{second}31.6
& 16.3 & \cellcolor{third}10.1 & 24.7
& \cellcolor{third}15.7 & \cellcolor{third}11.2 & 22.0 \\
w/o Alternating Estimation
& 10.1 & 7.7 & 14.1
& \cellcolor{best}17.5 & \cellcolor{best}12.2 & \cellcolor{best}18.5
& \cellcolor{second}12.4 & \cellcolor{best}9.2 & \cellcolor{second}17.2
& 22.1 & \cellcolor{second}15.8 & 33.9
& \cellcolor{second}15.2 & \cellcolor{second}9.6 & \cellcolor{second}22.1
& \cellcolor{second}15.5 & \cellcolor{second}10.9 & \cellcolor{second}21.1 \\
\midrule
w/o Specular
& \cellcolor{second}8.8 & \cellcolor{second}5.6 & \cellcolor{best}11.9
& 18.4 & 13.8 & 20.5
& 16.1 & 13.0 & 22.3
& \cellcolor{third}21.1 & 17.0 & \cellcolor{third}31.8
& \cellcolor{third}16.0 & 10.9 & \cellcolor{third}22.2
& 16.1 & 12.1 & \cellcolor{third}21.8 \\
w/o Diffuse
& 11.3 & 9.1 & 15.3
& \cellcolor{second}17.7 & \cellcolor{second}12.4 & \cellcolor{second}19.0
& \cellcolor{third}14.7 & \cellcolor{second}11.7 & \cellcolor{third}20.1
& 21.7 & 16.8 & 33.5
& 16.7 & 11.2 & 24.8
& 16.4 & 12.2 & 22.5 \\
\midrule
Ours
& \cellcolor{third}9.9 & 7.3 & \cellcolor{third}13.6
& \cellcolor{best}17.5 & \cellcolor{best}12.2 & \cellcolor{best}18.5
& \cellcolor{best}12.2 & \cellcolor{best}9.2 & \cellcolor{best}16.8
& \cellcolor{best}19.4 & \cellcolor{best}14.4 & \cellcolor{best}30.0
& \cellcolor{best}15.1 & \cellcolor{best}9.5 & \cellcolor{best}21.8
& \cellcolor{best}14.8 & \cellcolor{best}10.5 & \cellcolor{best}20.2 \\
\bottomrule
\end{tabular}

}
\end{table}


\begin{table}[t]
\caption{
\textbf{Novel view synthesis accuracy on synthetic data.} The results highlight that our proposed appearance model is essential for accurately recovering object appearance from sparse multi-view observations.
}
\label{tab:synth_nvs_quantitative}
\centering
\small
\setlength{\tabcolsep}{3.0pt}

\resizebox{\textwidth}{!}{%

\begin{tabular}{lcccccccccccc}
\toprule
\multirow{2}{*}{Method}
& \multicolumn{4}{c}{drain cleaner}
& \multicolumn{4}{c}{ottoman}
& \multicolumn{4}{c}{bunny} \\
\cmidrule(lr){2-5}\cmidrule(lr){6-9}\cmidrule(lr){10-13}
& PSNR $\uparrow$ & L1 $\downarrow$ & SSIM $\uparrow$ & LPIPS $\downarrow$
& PSNR $\uparrow$ & L1 $\downarrow$ & SSIM $\uparrow$ & LPIPS $\downarrow$
& PSNR $\uparrow$ & L1 $\downarrow$ & SSIM $\uparrow$ & LPIPS $\downarrow$ \\
\midrule
DG-Mesh~\cite{liu2024dynamic} & 35.23 & 0.00140 & 0.9921 & 0.0181 & 28.62 & 0.00515 & 0.9873 & \cellcolor{third}0.0186 & 36.26 & 0.00218 & 0.9895 & 0.0206 \\ \midrule
Single-Frame
& 40.03 & 0.00104 & 0.9947 & 0.0082
& 29.52 & 0.00454 & 0.9880 & \cellcolor{second}0.0148
& 39.42 & 0.00149 & 0.9931 & 0.0096 \\
w/o Motion-Aware Appearance
& 37.76 & 0.00209 & \cellcolor{best}0.9961 & \cellcolor{best}0.0059
& 29.45 & 0.00419 & 0.9890 & \cellcolor{second}0.0148
& 32.63 & 0.00388 & 0.9897 & 0.0097 \\
w/o Alternating Estimation
& \cellcolor{best}41.74 & \cellcolor{second}0.00096 & \cellcolor{best}0.9961 & \cellcolor{third}0.0066
& \cellcolor{best}32.88 & \cellcolor{best}0.00277 & \cellcolor{best}0.9904 & \cellcolor{best}0.0145
& \cellcolor{second}43.56 & \cellcolor{second}0.00094 & \cellcolor{best}0.9969 & \cellcolor{best}0.0062 \\
\midrule
w/o Specular
& 40.13 & \cellcolor{best}0.00093 & \cellcolor{third}0.9959 & 0.0067 
& 29.84 & 0.00406 & \cellcolor{third}0.9894 & \cellcolor{second}0.0148
& \cellcolor{third}40.93 & \cellcolor{third}0.00118 & \cellcolor{second}0.9963 & \cellcolor{third}0.0072 \\
w/o Diffuse
& \cellcolor{third}40.41 & 0.00124 & \cellcolor{second}0.9960 & 0.0067 
& \cellcolor{third}32.04 & \cellcolor{third}0.00292 & \cellcolor{second}0.9903 & \cellcolor{best}0.0145
& 38.86 & 0.00160 & \cellcolor{third}0.9950 & 0.0079 \\
\midrule
Ours
& \cellcolor{second}41.62 & \cellcolor{third}0.00098 & \cellcolor{best}0.9961 & \cellcolor{second}0.0065
& \cellcolor{second}32.77 & \cellcolor{second}0.00283 & \cellcolor{best}0.9904 & \cellcolor{best}0.0145
& \cellcolor{best}43.99 & \cellcolor{best}0.00091 & \cellcolor{best}0.9969 & \cellcolor{second}0.0063 \\
\bottomrule
\end{tabular}

}

\vspace{2mm}

\resizebox{\textwidth}{!}{%

\begin{tabular}{lcccccccccccc}
\toprule
\multirow{2}{*}{Method}
& \multicolumn{4}{c}{garden gnome}
& \multicolumn{4}{c}{drill}
& \multicolumn{4}{c}{Average} \\
\cmidrule(lr){2-5}\cmidrule(lr){6-9}\cmidrule(lr){10-13}
& PSNR $\uparrow$ & L1 $\downarrow$ & SSIM $\uparrow$ & LPIPS $\downarrow$
& PSNR $\uparrow$ & L1 $\downarrow$ & SSIM $\uparrow$ & LPIPS $\downarrow$
& PSNR $\uparrow$ & L1 $\downarrow$ & SSIM $\uparrow$ & LPIPS $\downarrow$ \\
\midrule
DG-Mesh~\cite{liu2024dynamic} & 34.86 & 0.00155 & 0.9918 & 0.0106 & 37.97 & 0.00120 & 0.9929 & 0.0130 & 34.59 & 0.00230 & 0.9907 & 0.0162 \\ \midrule
Single-Frame
& 36.94 & 0.00132 & 0.9933 & 0.0062
& 41.10 & 0.00092 & 0.9949 & 0.0065
& 37.40 & 0.00186 & 0.9928 & 0.0091 \\
w/o Motion-Aware Appearance
& 34.28 & 0.00207 & 0.9929 & 0.0064
& 42.18 & 0.00086 & 0.9968 & \cellcolor{best}0.0039
& 35.26 & 0.00262 & 0.9929 & 0.0081 \\
w/o Alternating Estimation
& \cellcolor{second}39.00 & \cellcolor{second}0.00110 & \cellcolor{third}0.9948 & \cellcolor{third}0.0061
& \cellcolor{best}46.32 & \cellcolor{best}0.00051 & \cellcolor{best}0.9973 & \cellcolor{third}0.0044
& \cellcolor{second}40.70 & \cellcolor{second}0.00126 & \cellcolor{second}0.9951 & \cellcolor{second}0.0076 \\
\midrule
w/o Specular
& \cellcolor{third}38.41 & \cellcolor{third}0.00115 & \cellcolor{second}0.9949 & \cellcolor{second}0.0055
& \cellcolor{third}45.25 & \cellcolor{second}0.00061 & \cellcolor{second}0.9972 & 0.0045
& \cellcolor{third}38.91 & 0.00159 & \cellcolor{third}0.9947 & \cellcolor{third}0.0077 \\
w/o Diffuse
& 37.45 & 0.00136 & 0.9943 & 0.0063
& 44.36 & \cellcolor{third}0.00065 & \cellcolor{third}0.9970 & 0.0046 
& 38.62 & \cellcolor{third}0.00155 & 0.9945 & 0.0080 \\ 
\midrule
Ours
& \cellcolor{best}39.86 & \cellcolor{best}0.00100 & \cellcolor{best}0.9953 & \cellcolor{best}0.0054
& \cellcolor{second}46.12 & \cellcolor{best}0.00051 & \cellcolor{best}0.9973 & \cellcolor{second}0.0043
& \cellcolor{best}40.87 & \cellcolor{best}0.00125 & \cellcolor{best}0.9952 & \cellcolor{best}0.0074 \\
\bottomrule
\end{tabular}

}
\end{table}

\begin{table}[t]
\caption{
  \textbf{Surface mesh reconstruction on synthetic data.} We evaluate geometric accuracy using Chamfer Distance (CD, $\downarrow$) and Normal Error ($\downarrow$). The results demonstrate the superior precision of our reconstruction compared to baseline methods. 
}
\label{tab:synth_mesh_quantitative}
\centering
\small
\setlength{\tabcolsep}{4pt}
\resizebox{\textwidth}{!}{%

\begin{tabular}{lcccccccccccc}
\toprule
\multirow{2}{*}{Method}
& \multicolumn{2}{c}{drain cleaner}
& \multicolumn{2}{c}{ottoman}
& \multicolumn{2}{c}{bunny}
& \multicolumn{2}{c}{garden gnome}
& \multicolumn{2}{c}{drill}
& \multicolumn{2}{c}{Average} \\
\cmidrule(lr){2-3}\cmidrule(lr){4-5}\cmidrule(lr){6-7}\cmidrule(lr){8-9}\cmidrule(lr){10-11}\cmidrule(lr){12-13}
& CD $\downarrow$ & Normal $\downarrow$
& CD $\downarrow$ & Normal $\downarrow$
& CD $\downarrow$ & Normal $\downarrow$
& CD $\downarrow$ & Normal $\downarrow$
& CD $\downarrow$ & Normal $\downarrow$
& CD $\downarrow$ & Normal $\downarrow$ \\
\midrule
DG-Mesh~\cite{liu2024dynamic}
& 0.045
& 33.78
& 0.066
& 30.42
& 0.036
& 39.12
& 0.027
& 30.67
& 0.032
& 29.71
& 0.041
& 32.74 \\
\midrule
Single-Frame
& 0.051
& 30.48
& \cellcolor{second}0.030
& 27.92
& 0.036
& 31.07
& \cellcolor{second}0.021
& 32.75
& 0.028
& 33.71
& 0.033
& 31.19 \\
w/o Motion-Aware Appearance
& \cellcolor{second}0.013
& \cellcolor{best}12.79
& 0.042
& 31.75
& 0.035
& 26.41
& 0.038
& 36.24
& \cellcolor{second}0.014
& \cellcolor{second}18.75
& 0.028
& 25.19 \\
w/o Alternating Estimation
& \cellcolor{best}0.012
& \cellcolor{third}13.90
& \cellcolor{best}0.028
& \cellcolor{best}19.77
& \cellcolor{second}0.024
& \cellcolor{best}18.99
& \cellcolor{third}0.024
& \cellcolor{second}28.45
& 0.016
& 22.32
& \cellcolor{second}0.021
& \cellcolor{second}20.69 \\
\midrule
w/o Specular
& \cellcolor{best}0.012
& 14.83
& 0.034
& 26.99
& \cellcolor{best}0.023
& \cellcolor{third}20.98
& \cellcolor{second}0.021
& \cellcolor{third}28.93
& \cellcolor{best}0.013
& \cellcolor{third}19.89 
& \cellcolor{second}0.021
& \cellcolor{third}22.32 \\
w/o Diffuse
& \cellcolor{third}0.014
& 14.74
& \cellcolor{second}0.030
& \cellcolor{third}25.76
& 0.033
& 22.10
& 0.029
& 33.03
& \cellcolor{third}0.015
& 21.59 
& \cellcolor{third}0.024
& 23.44 \\
\midrule
Ours
& \cellcolor{second}0.013
& \cellcolor{second}13.42
& \cellcolor{third}0.032
& \cellcolor{second}24.33
& \cellcolor{third}0.026
& \cellcolor{second}20.31
& \cellcolor{best}0.016
& \cellcolor{best}25.45
& \cellcolor{best}0.013
& \cellcolor{best}18.72
& \cellcolor{best}0.020
& \cellcolor{best}20.45 \\
\bottomrule
\end{tabular}

}
\end{table}

\begin{figure*}[t]
  \centering
  \includegraphics[width=\linewidth]{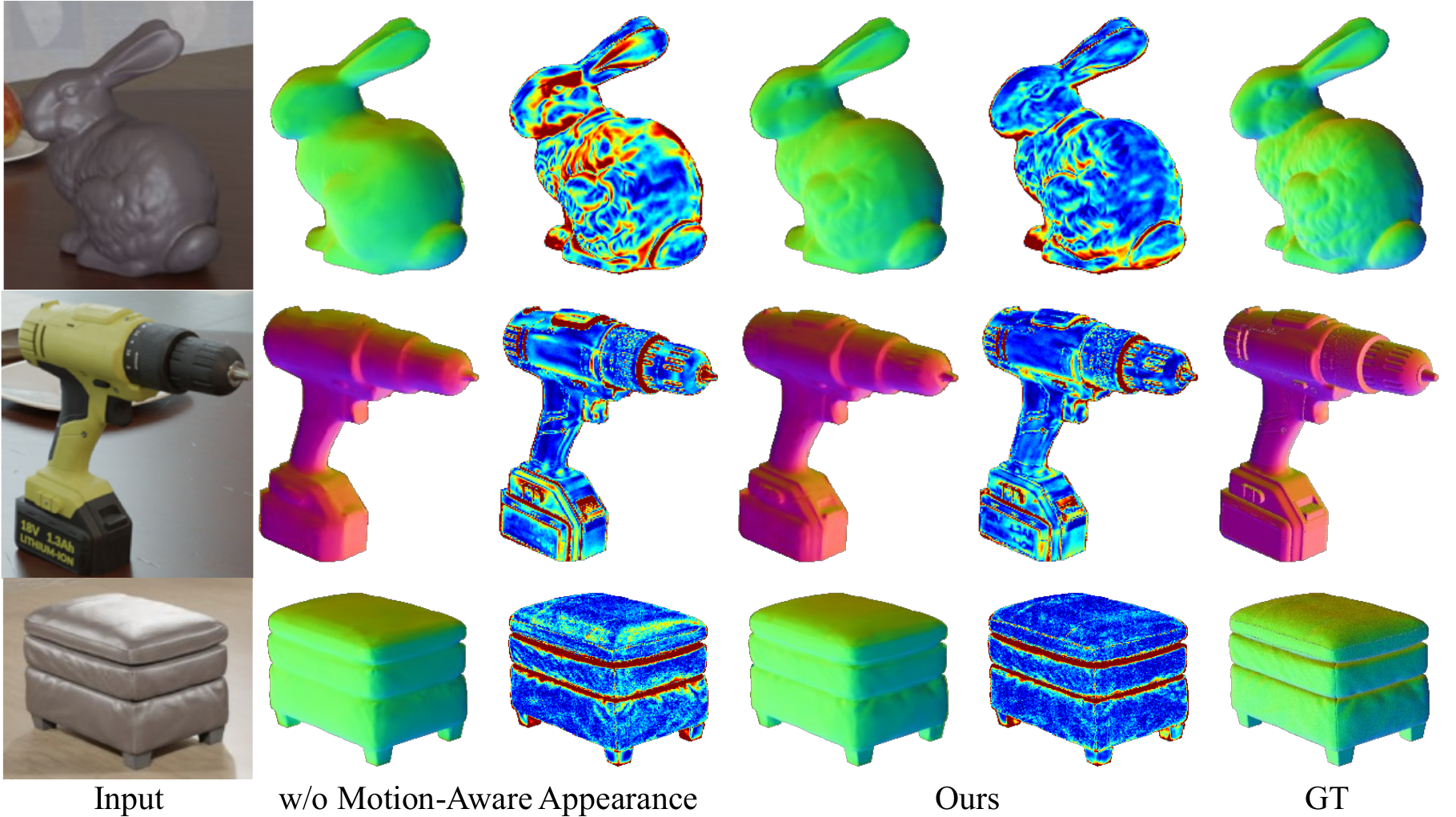}
  \caption{
\textbf{Visualization of recovered surface normals.} The error maps on the right visualize per-pixel estimation discrepancies. These results demonstrate that proper appearance modeling is essential for effectively leveraging radiometric cues to reconstruct fine surface details. 
  }
  \label{fig:results_normal}
\end{figure*}

\begin{figure*}[t]
  \centering
  \includegraphics[width=\linewidth]{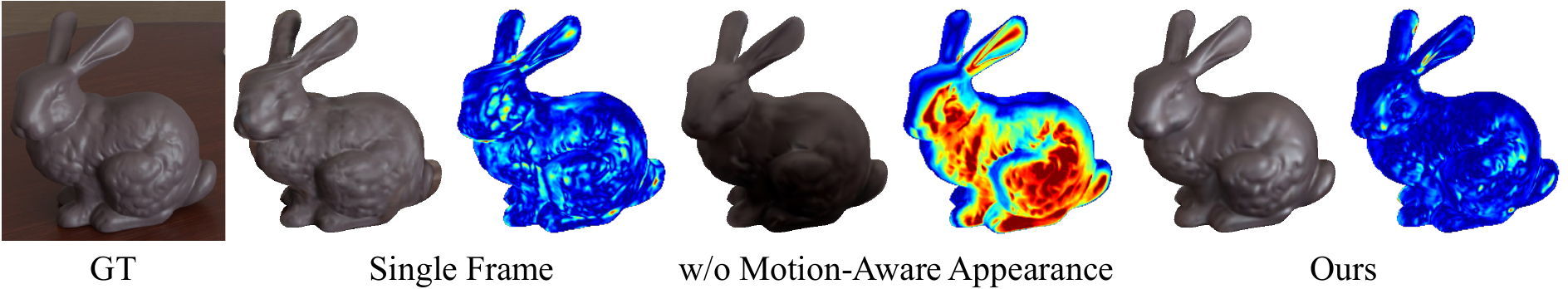}
  \caption{
\textbf{Novel view Synthesis.} \kohei{Our appearance model significantly improves the fidelity of novel view synthesis when applied to objects with specularity.}
  }
  \label{fig:results_appearance}
\end{figure*}

We validate the effectiveness of our method using a newly created synthetic multi-view video dataset and established real-world datasets.

\vspace{-8pt}
\paragraph{\textbf{Implementation Details.}}
We implement our framework using a single NVIDIA RTX A6000 GPU (48 GB VRAM). On average, pose estimation and Gaussian refinement require 3 and 4 minutes per frame, respectively, with a total optimization time of approximately 7 hours for a typical sequence. In practice, this can be sped up, \eg, by keyframing during the alternating estimation.

\vspace{-8pt}
\paragraph{\textbf{Datasets.}} 
For quantitative evaluation against ground-truth geometry, we generated a synthetic dataset consisting of 5 distinct sequences, each 29 frames in length. These sequences feature 5 different moving objects within 3 varied environments. Each sequence provides four-view videos with corresponding ground-truth 3D meshes.
We further evaluate our method on two real world datasets capturing diverse human-object interactions: the HODome dataset~\cite{zhang2023neuraldome} (\ryosuke{high-fidelity capture of human-object interaction})
and the HO3D dataset~\cite{hampali2020honnotate} (hand-object interaction focused on object pose estimation).

\vspace{-8pt}
\paragraph{\textbf{Baseline Methods.}}
We evaluate the effectiveness of our method by comparing it with its ablated variants. ``Single-Frame'' is our initialization process and recovers geometry and appearance from only the initial timestep frames similar to MAtCha~\cite{guedon2025matcha}. ``w/o Motion-Aware Appearance'' skips the optimization of geometry and appearance with the proposed appearance model. ``w/o Alternating Estimation'' skips the pose estimation and optimizes all the parameters from the beginning of the estimation at each timestep. For more detailed analysis on the appearance model, we also compare our method with ``w/o Specular'' and ``w/o Diffuse'' which ablate the specular and diffuse terms in \cref{eq:appearance_model}, respectively. \ryosuke{Note that for ``w/o Diffuse'', we still use 0th order SH coefficients, which represents view- and motion-invariant color to let the gaussians have individual color.}

\vspace{-8pt}
\paragraph{\textbf{Evaluation Metrics.}}
To assess the quality of the recovered surface details, we compute the angular error between normal maps rendered from the Gaussians and the ground truth. We report the mean, median, and 80th percentile errors.
Novel view synthesis accuracy is measured using PSNR, L1 error, SSIM, and LPIPS, calculated exclusively within the object regions.
For the reconstructed mesh, we evaluate geometric fidelity using the Chamfer Distance (CD) 
and the average angular error of the surface normals. 
\ryosuke{To calculate the latter, we compare the normal of each sampled point on the extracted mesh surface with the normal of the nearest point on the ground-truth surface.}

\subsection{Evaluation on Synthetic Data}

We first quantitatively evaluate our method using synthetic data with ground-truth surface details. \Cref{tab:synth_normal_quantitative,tab:synth_nvs_quantitative,tab:synth_mesh_quantitative} summarize the accuracy of the recovered surface normals, novel view synthesis, and 3D mesh models. These results demonstrate that our proposed components are essential for reconstructing detailed geometry and appearance. Notably, our method achieves significantly higher accuracy than the single-frame baselines in surface normal estimation and novel view synthesis for the bunny and drill sequences, which feature complex geometries that pose a challenge for the baseline.

Qualitative results for recovered surface normals \kohei{and novel view synthesis} are shown in \cref{fig:results_normal,fig:results_appearance}. Without effectively leveraging radiometric cues, learned priors and cross-view photometric consistency provide insufficient information to resolve fine surface details. Consequently, the ``w/o Motion-Aware Appearance'' baseline recovers only coarse geometry \kohei{and appearance}. In contrast, our method successfully reconstructs accurate surface details \kohei{and novel views} by fully exploiting radiometric cues through the proposed appearance model.

In \cref{tab:synth_normal_quantitative,tab:synth_nvs_quantitative,tab:synth_mesh_quantitative}, we compare our method with DG-Mesh~\cite{liu2024dynamic}, a representative baseline for surface mesh recovery from multi-view RGB video. While DG-Mesh supports deformable objects, it requires extensive camera motion to jointly estimate geometry, appearance, and deformation. Consequently, it struggles with the sparse inputs provided by static cameras. In contrast, our method effectively leverages sparse, static-view observations by jointly optimizing object pose, geometry, and appearance. 
Note that we only report results of DG-Mesh~\cite{liu2024dynamic} on the synthetic data as it did not produce meaningful results on real data due to occlusions from human subjects and large motions.

\begin{figure*}[t]
  \centering
  \includegraphics[width=\linewidth]{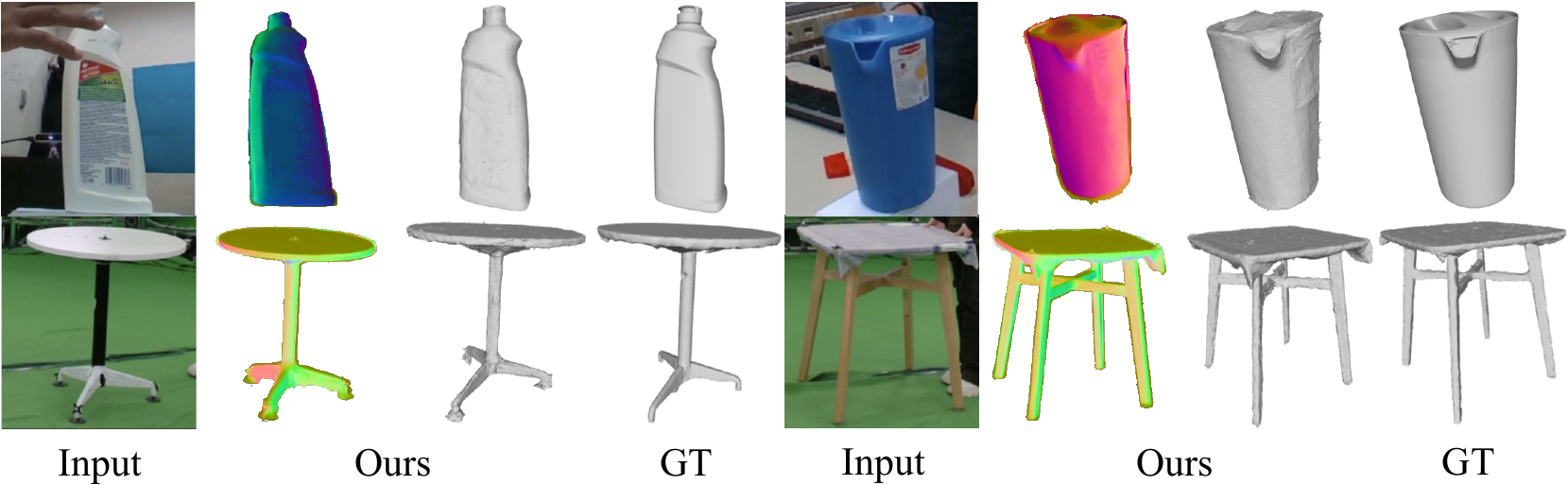}
  \caption{
\textbf{Surface normals and mesh models recovered from real-world datasets.} Qualitative results on HO3D~\cite{hampali2020honnotate} and HODome~\cite{zhang2023neuraldome}
demonstrate the effectiveness and robustness of our method across diverse human-object interaction scenarios.
  }
  \label{fig:results_geometry_real}
\end{figure*}

\begin{figure*}[t]
  \centering
  \includegraphics[width=\linewidth]{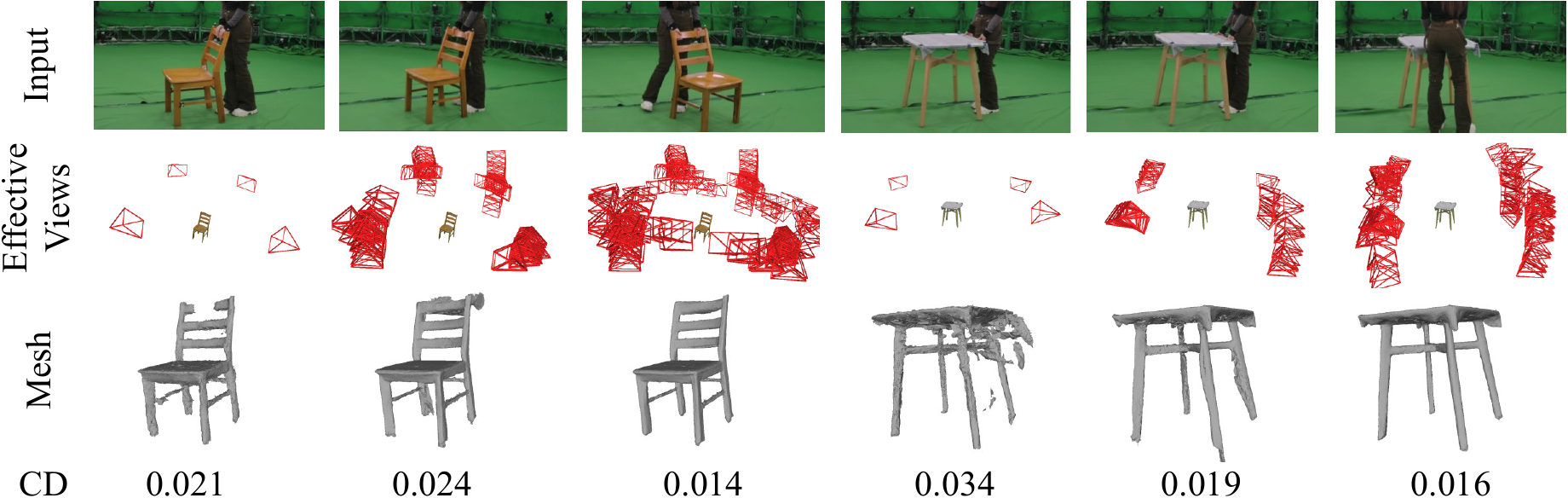}
  \caption{
\textbf{Visualization of alternating estimation progress.} We show the recovered geometry and object poses (represented as effective camera views) at several timesteps. As the temporal window expands and the effective camera coverage becomes more informative, the geometric accuracy of the reconstruction progressively improves.
  }
  \label{fig:results_alternating_estimation}
\end{figure*}

\begin{table}[t]
\caption{
\textbf{Quantitative results on real-world datasets.}
Our method achieves better geometry accuracy and novel view synthesis quality than the baselines.
}
\label{tab:result_real}
\centering
\small
\setlength{\tabcolsep}{3.5pt}
\resizebox{\textwidth}{!}{%

\begin{tabular}{lcccccccccc}
\toprule
\multirow{2}{*}{Method}
& \multicolumn{5}{c}{HO3D}
& \multicolumn{5}{c}{HODome} \\
\cmidrule(lr){2-6}\cmidrule(lr){7-11}
& CD $\downarrow$ & PSNR $\uparrow$ & L1 $\downarrow$ & SSIM $\uparrow$ & LPIPS $\downarrow$
& CD $\downarrow$ & PSNR $\uparrow$ & L1 $\downarrow$ & SSIM $\uparrow$ & LPIPS $\downarrow$ \\
\midrule
Single-Frame
& \cellcolor{second}0.011 & \cellcolor{second}27.60 & \cellcolor{best}0.01017 & \cellcolor{third}0.9565 & \cellcolor{third}0.0412
& \cellcolor{third}0.024 & \cellcolor{second}39.00 & \cellcolor{second}0.00101 & \cellcolor{third}0.9946 & \cellcolor{third}0.0062 \\
w/o Motion-Aware Appearance
& \cellcolor{best}0.004 & \cellcolor{third}27.09 & \cellcolor{second}0.01344 & \cellcolor{second}0.9580 & \cellcolor{second}0.0345
& \cellcolor{second}0.017 & \cellcolor{third}38.04 & \cellcolor{third}0.00127 & \cellcolor{second}0.9951 & \cellcolor{second}0.0057 \\
\midrule
Ours
& \cellcolor{best}0.004 & \cellcolor{best}28.11 & \cellcolor{best}0.01017 & \cellcolor{best}0.9625 & \cellcolor{best}0.0296
& \cellcolor{best}0.016 & \cellcolor{best}39.71 & \cellcolor{best}0.00099 & \cellcolor{best}0.9957 & \cellcolor{best}0.0056 \\
\bottomrule
\end{tabular}

}
\end{table}

\subsection{Evaluation on Real-World Data}
We further evaluate our method on several real-world datasets. To maintain our assumption of a single rigidly moving object, we pre-process the scenes to remove human subjects from the representation. During initialization, we exclude Gaussians projected onto annotated human segmentation masks; during multi-frame optimization, we mask human regions when computing loss functions.

\Cref{fig:results_geometry_real,tab:result_real} present qualitative and quantitative results for surface normals, 3D mesh models, and novel view synthesis. While these datasets provide ground-truth meshes, they often lack fine geometric details; therefore, we restrict our geometric evaluation to the Chamfer Distance (CD). 
Although real-world conditions may not strictly adhere to all model assumptions, our alternating estimation framework and appearance model consistently improve reconstruction quality, demonstrating the robustness of our approach.

\Cref{fig:results_alternating_estimation} visualizes the recovered 3D shapes and object poses at several timesteps of the alternating optimization process. We represent object poses as ``effective camera views''---the camera's position relative to the object's local coordinate system. The results confirm that our method successfully leverages object motion, progressively improving geometric accuracy as the diversity of effective viewpoints increases.


\section{Conclusion}

We introduced a method to recover the 3D shape, appearance, and time-varying poses of a rigidly moving object from static, sparse-view RGB videos. By introducing an alternating estimation framework that leverages opportunistic motion and a dedicated model for complex view and pose-dependent appearance, our method successfully reconstructs detailed geometry and appearance from limited viewpoints. We believe this approach serves as a practical tool for 3D scene understanding in real-world scenarios, such as home safety monitoring for the elderly or children.

Despite its effectiveness in handling sparse observations and dynamic appearance, certain limitations remain. First, because our method initializes canonical Gaussians using learned priors, it may fail on uncommon objects not represented in the training data. Second, our model assumes a simplified surface reflection model---comprising Lambertian and uniform, reflection-centered specular components---and distant illumination. Consequently, it may struggle with complex effects such as Fresnel reflection, microfacet-based light transport, or near-field lighting. Factorizing the specular component to disentangle the illumination for relighting represents another promising research direction. Finally, our framework assumes rigid bodies and does not currently handle deformable components. We plan to address these challenges in future work by using our current estimation framework as a foundation for more sophisticated modeling.

\section*{Acknowledgement}

This work was in part supported by
JSPS KAKENHI 
21H04893, JST JPMJAP2305, and
the European Union (ERC Advanced Grant explorer Funding ID \#101097259) .


%
%

\bibliographystyle{splncs04}
\bibliography{cleaned_main}

\renewcommand{\thesection}{\Alph{section}}
\setcounter{section}{0}

\section{Results on In-the-wild Capture}

\begin{figure*}[t]
  \centering
  \includegraphics[width=\linewidth]{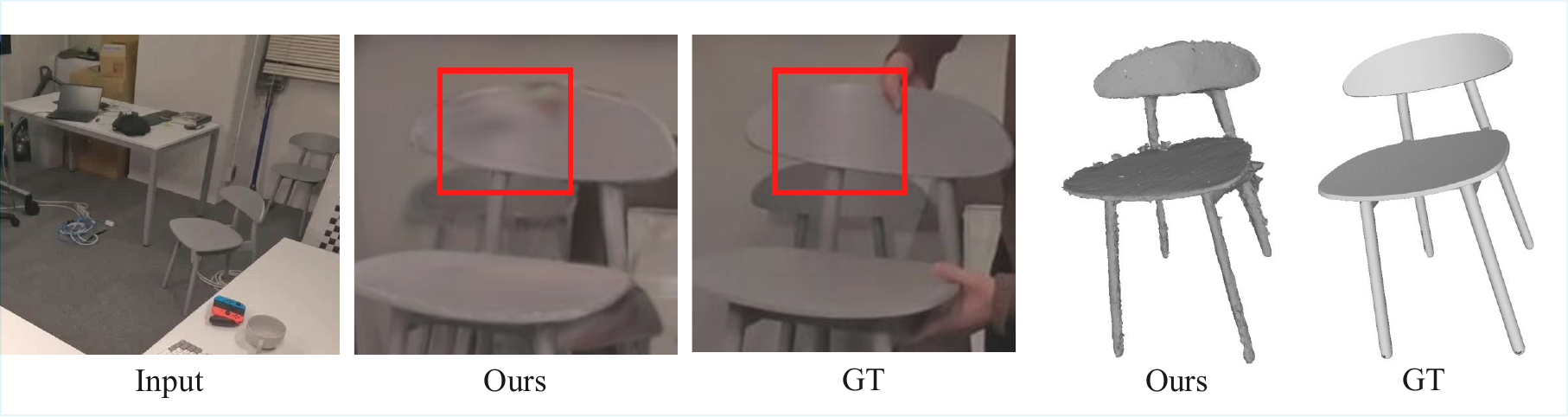}
  \caption{
\textbf{Results on in-the-wild capture.} \ryosuke{Our method reconstructs faithful appearance and geometry from real-world capture, including specular highlight on the object.}
  }
  \label{fig:self_captured_data_results_supp}
\end{figure*}

In \cref{fig:self_captured_data_results_supp} and the supplementary video,
we test our method on in-the-wild videos captured by ourselves.
We place cameras at the four corners of a room and record human--object interactions.
To obtain ground-truth geometry, we also scan each object using a 3D scanner.
\ryosuke{Our method reconstructs accurate object geometry and appearance including specular highlight on the object even from these in-the-wild videos. The results show promise for great applicability of our method to real-world scenarios such as home safety monitoring for elderly people or children.}

\section{Implementation Details}

The learnable parameters of our method are the parameters of 2D Gaussians, the appearance parameters shared across the object ($\theta_s$ and $\theta_d$), and per-frame object poses (quaternion $\mathbf{q}_\mathrm{obj}^t$ and translation $\mathbf{t}_\mathrm{obj}^t$ for each timestep). The parameters of each Gaussian are its 3D position $\boldsymbol{\mu}_i$, rotation (quaternion) $\boldsymbol{q}_i$, 2D scale $\boldsymbol{s}_i$, opacity $\sigma_i$, segmentation mask value $m_i$, and diffuse albedo $\mathbf{k}_{d,i}$. During single-view 3D reconstruction and alternating estimation, we optimize a single set of SH coefficients $\theta_i$ for each Gaussian instead of $\mathbf{k}_{d,i}$, $\theta_s$, and $\theta_d$. The sets of SH coeeficients $\theta_s$, $\theta_d$, and $\theta_i$ correspond to spherical harmonics of orders 9, 3, and 3, respectively.
We use the Adam optimizer for all the training stages. 

\paragraph{\textbf{Single-Frame 3D Reconstruction}} We first optimize the parameters of the Gaussians except for $m_i$ using the multi-view images at the first timestep $\{\mathbf{I}_v^1|v=1, \dots, V\}$. The training loss for this single-frame reconstruction is the same as that used in the free Gaussians refinement stage of MAtCha Gaussians~\cite{guedon2025matcha}. 
We use a weighted sum of a rendering loss $\mathcal{L}_\mathrm{RGB}$, a depth-normal consistency loss $\mathcal{L}_\mathrm{n}$, a depth prior loss $\mathcal{L}_\mathrm{pdepth}$, a normal prior loss $\mathcal{L}_\mathrm{pnormal}$, and a curvature prior loss $\mathcal{L}_\mathrm{pcurv}$.
Since a depth distortion loss~\cite{Huang2DGS2024} is disabled in MAtCha Gaussians codebase, we follow them and set it to zero.
The number of iterations for this reconstruction is 7000, which takes approximately 7 minutes.

We then optimize the mask values $\{m_i\}$ using 2D segmentation masks $\{\mathrm{M}_v^1|v=1, \dots, V\}$ generated via SAM2~\cite{ravi2024sam2}. We use an image-space loss composed of L1 and D-SSIM terms to evaluate the discrepancies between the rendered and pseudo ground-truth masks. The number of iterations and the running time are 7000 and 7 minutes, respectively.

\paragraph{\textbf{Pose Estimation}} 
Given a new frame (the $k$-th frame), we first optimize the corresponding object pose $\mathbf{q}_\mathrm{obj}^k$ and $\mathbf{t}_\mathrm{obj}^k$ using the multi-view images $\{\mathbf{I}_v^k|v=1, \dots, V\}$. We initialize the object pose with the estimated pose of the previous frame (\ie, $\mathbf{q}_\mathrm{obj}^{k-1}$ and $\mathbf{t}_\mathrm{obj}^{k-1}$). We set $\lambda_\mathrm{RGB}$ and $\lambda_\mathrm{entropy}$ to 1. For only one sequence from HODome dataset, we set $\lambda_\mathrm{entropy}$ to 2 to ensure strong regularization. The number of iterations for each frame is 1000. For each iteration, we randomly sample one input viewpoint $v_s\in\{1,\dots,V\}$ and evaluate the training loss using image $I_{v_s}^k$. The running time is approximately 3 minutes per frame.

\paragraph{\textbf{Gaussian Refinement}}
We then refine all learnable parameters using all processed frames $\{\mathbf{I}_v^t|v=1, \dots, V, t=1, \dots, k\}$. We set $\lambda_\mathrm{RGB}$, $\lambda_\mathrm{n}$, $\lambda_\mathrm{pnormal}$, $\lambda_\mathrm{pdepth}$, and $\lambda_\mathrm{entropy}$ to 1, 0.05, 0.25, 0.375, and 1.0, respectively. We apply adaptive density control~\cite{kerbl3Dgaussians} to the Gaussians every 100 iterations and reset the opacity $\sigma_i$ to 0.01 every 1000 iterations except for the last 200 iterations. For each iteration, we randomly sample one input viewpoint and one timestep, and evaluate the loss using the corresponding image. The number of iterations and running time for each iteration are 2000 and 4 minutes, respectively.
\ryosuke{After alternating estimation for all the time steps is done, we apply an additional refinement for 7000 iterations by reducing the geometric prior regularization, $\mathcal{L}_\mathrm{pnormal}$ and $\mathcal{L}_\mathrm{pdepth}$, since we already have enough images for supervision.}

\paragraph{\textbf{Refinement with Motion-Aware Appearance Modeling}}
Once the alternating estimation is complete, we finally refine all learnable parameters using all frames $\{\mathbf{I}_v^t|v=1, \dots, V, t=1, \dots, T\}$. In this refinement, we replace the per-Gaussian SH coefficients $\{\theta_i\}$ with per-Gaussian diffuse albedo $\{\mathbf{k}_{d,i}\}$ and the shared SH coefficients $\theta_s$ and $\theta_d$. We initialize the diffuse albedo $\mathbf{k}_{d,i}$ so that it corresponds to the 0-th order component of $\theta_i$. We initialize the SH coefficients so that the corresponding spherical functions $\mathbf{f}(\mathbf{v}; \theta_s)$ and $\mathbf{f}(\mathbf{v}; \theta_d)$ always output 0 and a positive constant value, respectively. We set $\lambda_\mathrm{RGB}$, $\lambda_\mathrm{normal}$ and $\lambda_\mathrm{entropy}$ to 1, 0.05 and 1, respectively. We use the adaptive density control every 100 iterations and the opacity reset every 1000 iterations for the first 3500 iterations. The number of iterations for this final refinement is 12500, which takes approximately 2 hours.

\paragraph{\textbf{Meshing}}
For each Gaussian, we place pivot points at its center $\boldsymbol{\mu}_i$ and at the eight vertices of a cuboid aligned with the Gaussian orientation $\mathbf{q}_i$.
The side lengths of the cuboid are set to three times the Gaussian scale $\mathbf{s}_i$.
Along the surface normal direction, we use a fixed length of $2\times10^{-4}$.
For each pivot point, we use an SDF value computed by TSDF fusion as an initial value for the optimization. This post-processing takes approximately 10 minutes.

\section{Dataset Details}

\subsection{Synthetic Dataset}

The synthetic dataset consists of multi-view videos of five object-environment pairs drawn from five objects and three surrounding environments, namely, a drain cleaner in a room, an ottoman in a room, a bunny on a table in a room, a garden gnome on a table in a room, and a drill on a table in an airport. 
In the room environment, the scene consists of 3D models of furnitures, walls, and a floor illuminated by point light sources, whereas the other environments are composed of several 3D furniture models \ryosuke{and an environment map}. 
In each sequence, the object rotates in place while the surrounding environment remains static. 
The rotation between adjacent frames is 12.9 degrees.
Four training viewpoints are placed around the object (corresponding to the four corners of the room in the room environment), and three intermediate viewpoints between them are used as test views for novel view synthesis. 
All images are rendered using Blender Cycles, producing photorealistic renderings with global illumination and surface roughness effects.

\subsection{HO3D Dataset~\cite{hampali2020honnotate}}

For the HO3D dataset (a hand-object interaction dataset), we use the sequences that provide multi-view images, namely AP1, GPMF1, MDF1, and SB1. 
These sequences contain a variety of objects such as a pitcher, a bleach cleaner bottle, and a potted meat can. 
Among the five available camera views, we use four views surrounding the object as training views and use one remaining view for novel view synthesis evaluation. 
Since the object motion in HO3D is relatively slow, we subsample the input video by taking every 5th to 15th frame \ryosuke{so that the total number of frames becomes approximately 100} and use the resulting frame sequence as input to our method.

\subsection{HODome Dataset~\cite{zhang2023neuraldome}}
For the HODome dataset (high-fidelity capture of human-object interaction), we use the box, chair, desk, table, talltable, trashcan, and trolleycase sequences, which cover a diverse set of objects. 
We select four views surrounding the object as training views and seven views from the remaining views for novel view synthesis evaluation. 
Since the sequences are temporally dense (60 FPS), \ryosuke{we subsample them so that the total number of frames becomes around 50 before feeding them into our method. Note that we use only 33 frames for talltable sequence since the motion is relatively subtle.}

\section{Additional Experimental Results}

\subsection{Additional Qualitative Results on Synthetic Data}

\begin{figure*}[t]
  \centering
  \includegraphics[width=\linewidth]{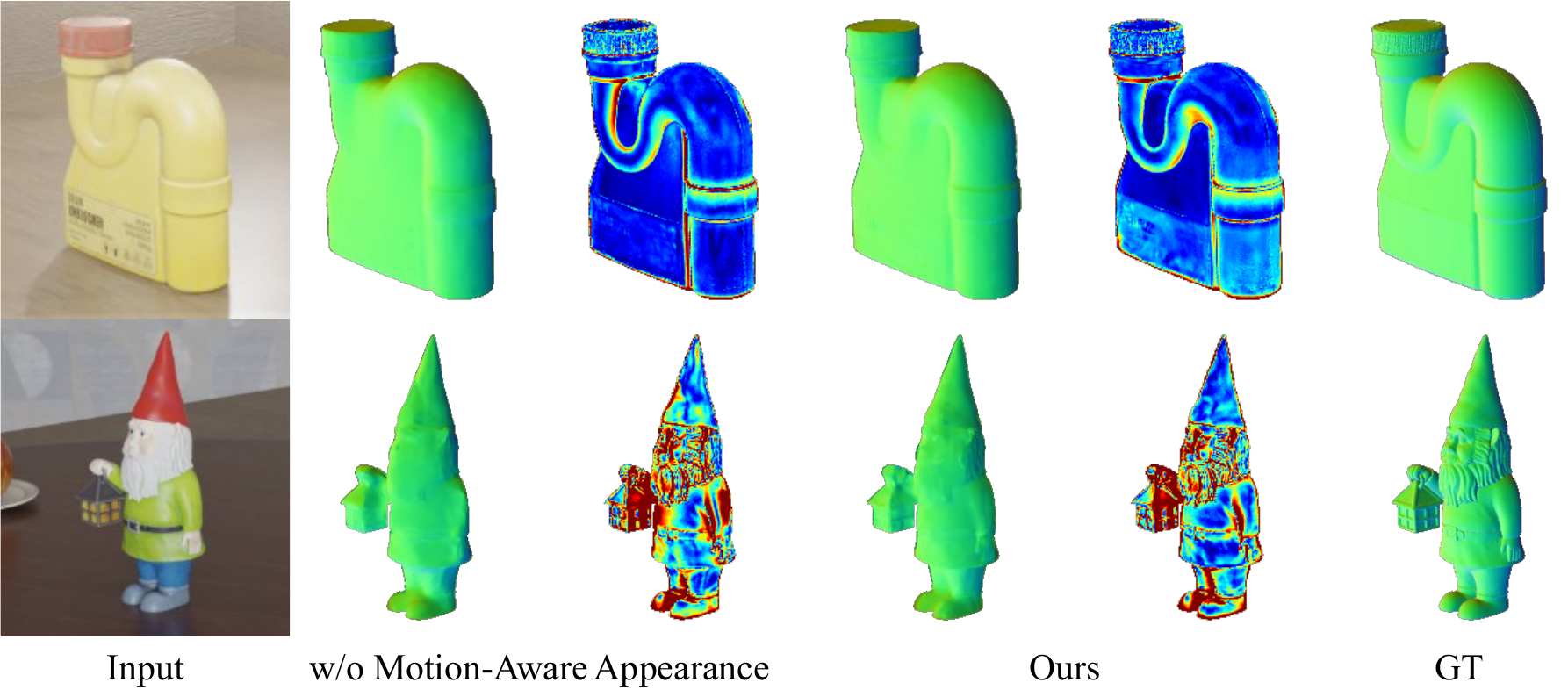}
  \caption{
\textbf{Additional Visualization of Recovered Surface Normals.} Our recovered surface normals are at least comparable to those of the baselines, suggesting the robustness of our method.
  }
  \label{fig:results_normal_supp}
\end{figure*}

\begin{figure*}[t]
  \centering
  \includegraphics[width=\linewidth]{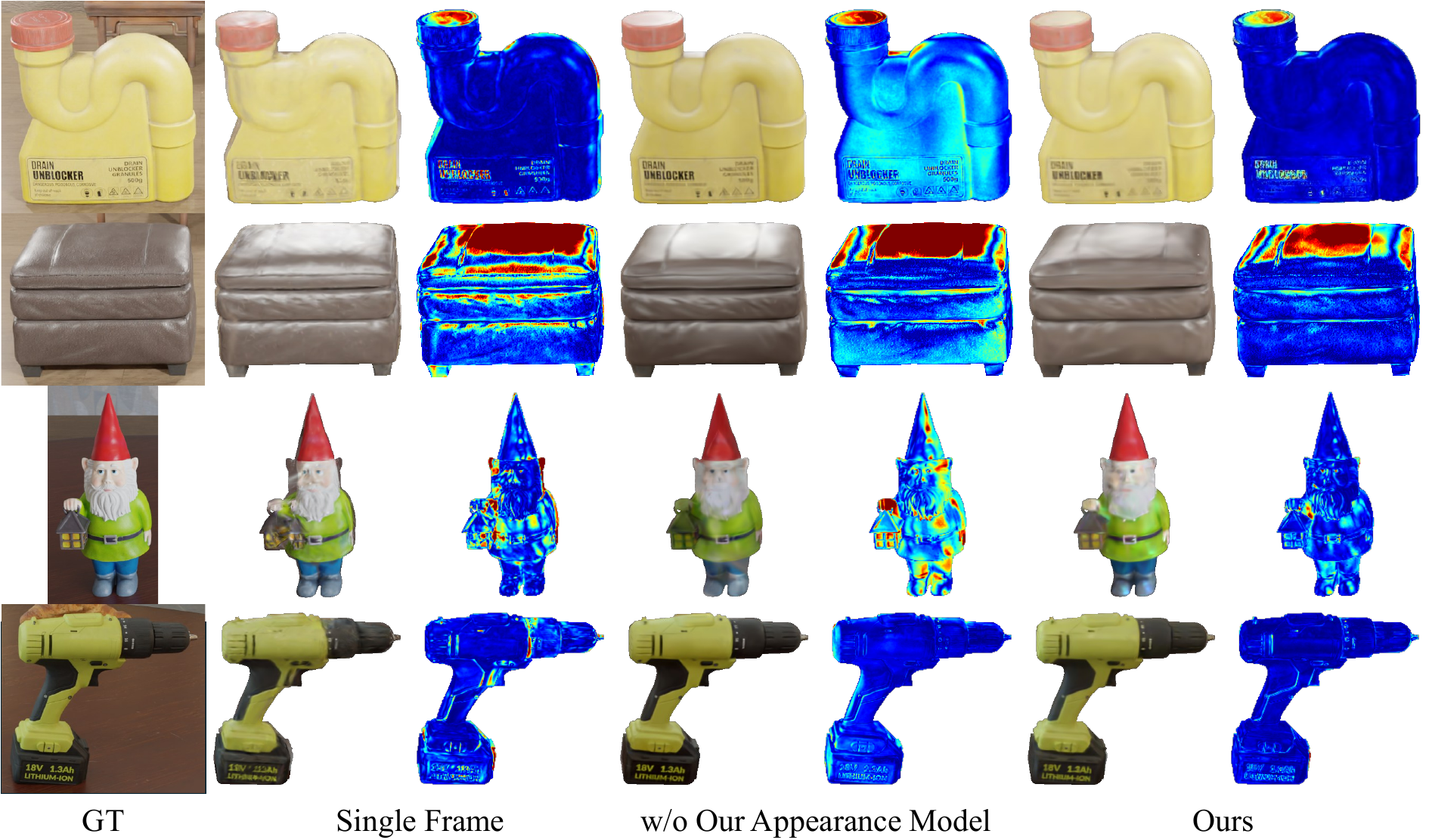}
  \caption{
\textbf{Additional Novel View Synthesis Results.} Our appearance model improves the fidelity of novel view synthesis on different objects. The ottoman (the second row) remains challenging due to its limited surface-normal variation.
  }
  \label{fig:results_appearance_supp}
\end{figure*}

\begin{figure*}[t]
  \centering
  \includegraphics[width=\linewidth]{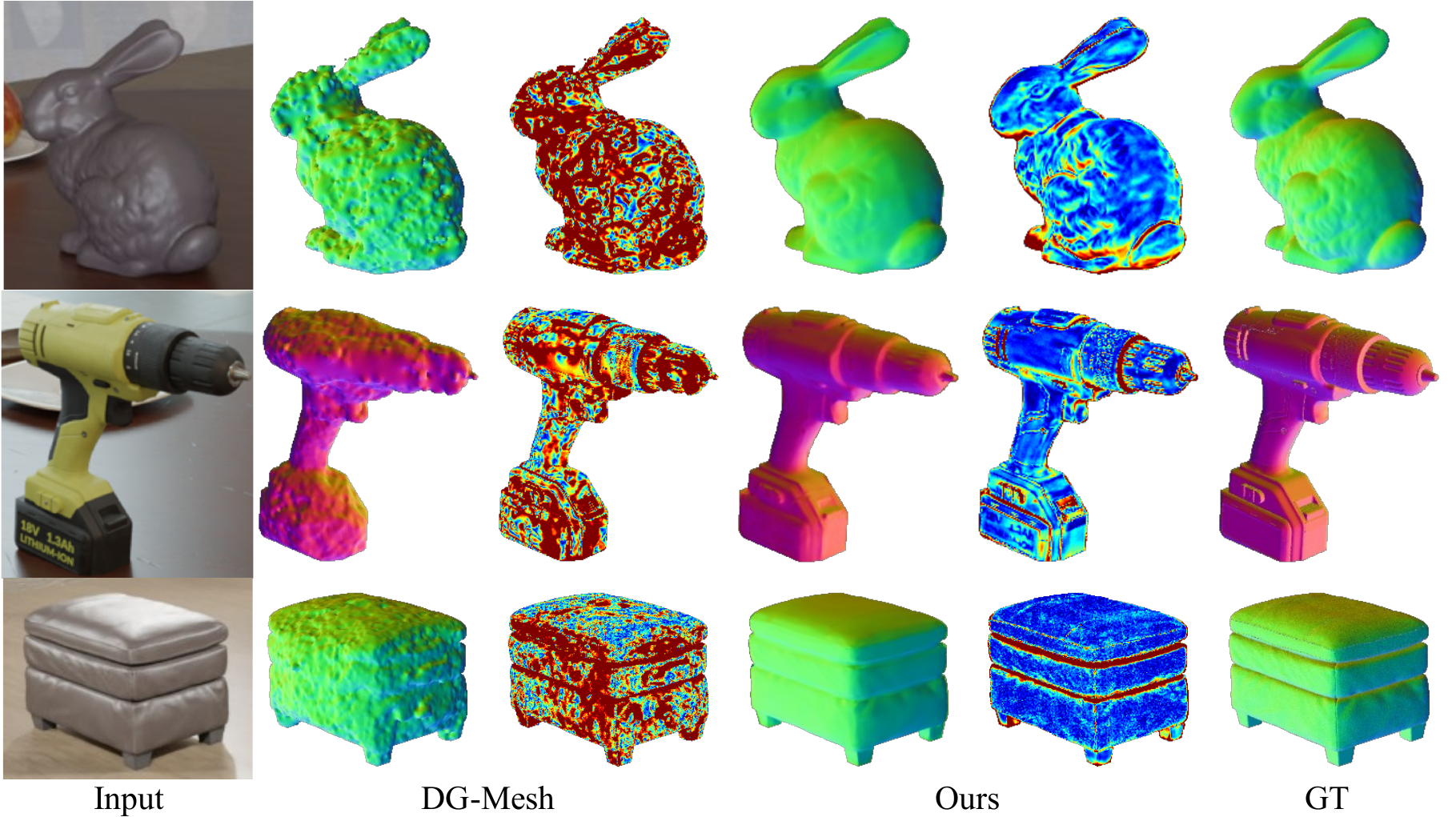}
  \caption{
\textbf{Qualitative comparison with DG-Mesh~\cite{liu2024dynamic}.} While DG-Mesh supports deformable objects, it struggles with the sparse inputs provided by static cameras. 
  }
  \label{fig:results_comparison_dgmesh}
\end{figure*}

\begin{figure*}[t]
  \centering
  \includegraphics[width=\linewidth]{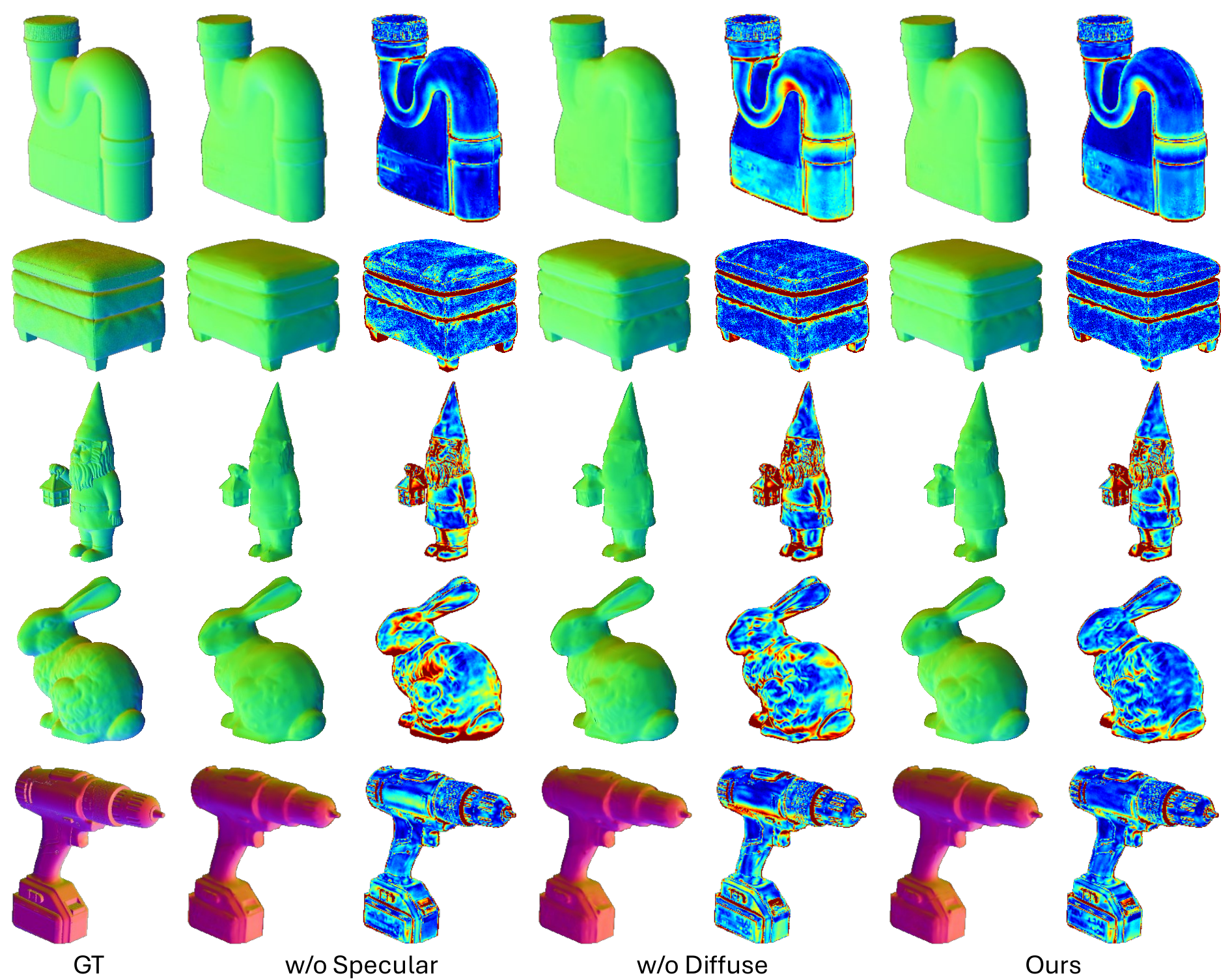}
  \caption{
\textbf{Effectiveness of diffuse and specular components for surface normal recovery.}
Modeling both diffuse and specular components is critical for recovering accurate surface normals on objects with diverse material properties, where view-dependent effects can otherwise bias the reconstruction.
  }
  \label{fig:results_ablation_diffuse_specular_normal}
\end{figure*}

\begin{figure*}[t]
  \centering
  \includegraphics[width=\linewidth]{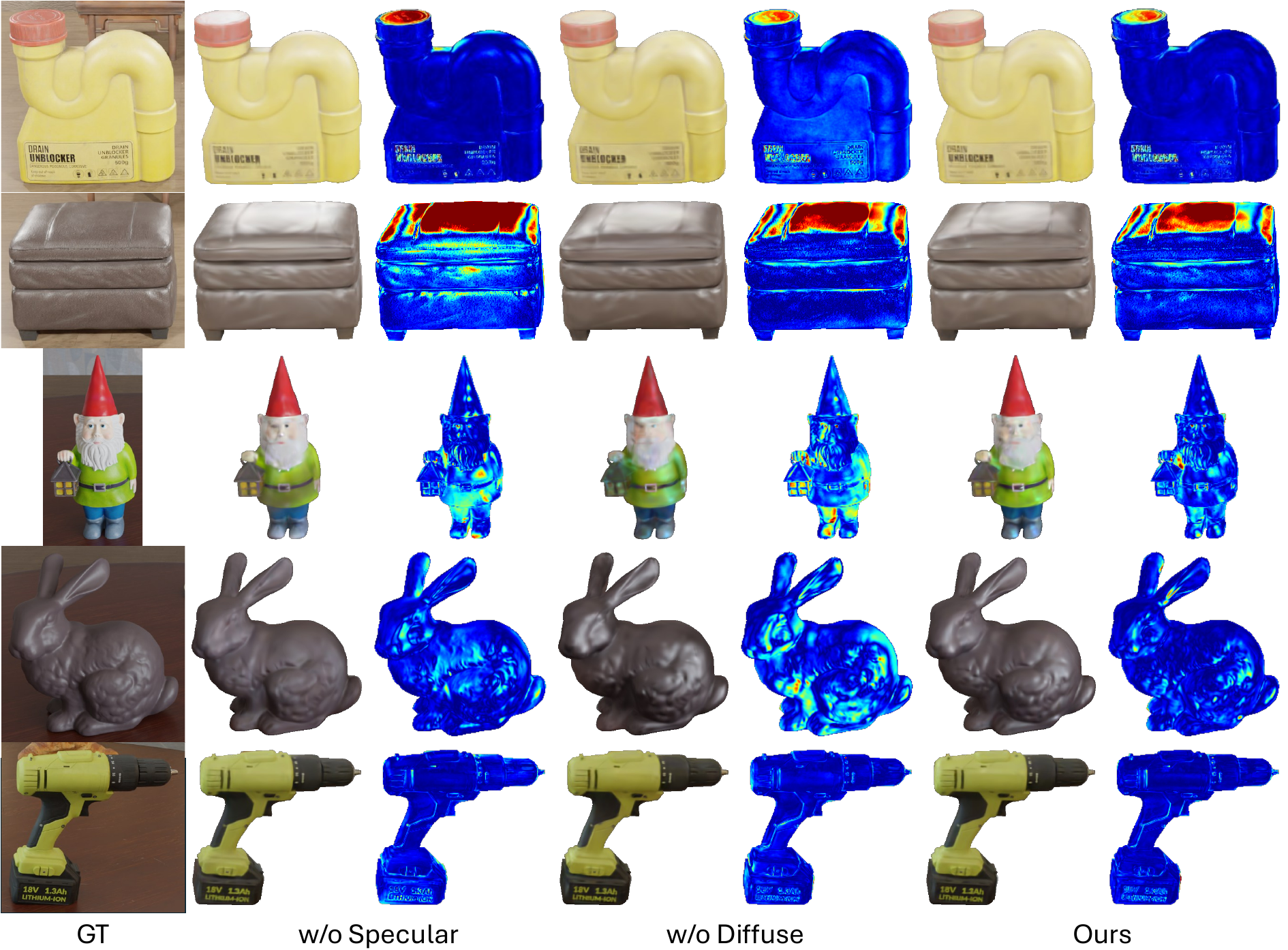}
  \caption{
    \textbf{Effectiveness of diffuse and specular components for novel view synthesis.}
Our full model can faithfully reproduce both specular highlights and diffuse shading.
  }
  \label{fig:results_ablation_diffuse_specular_appearance}
\end{figure*}

\begin{figure*}[t]
  \centering
  \includegraphics[width=\linewidth]{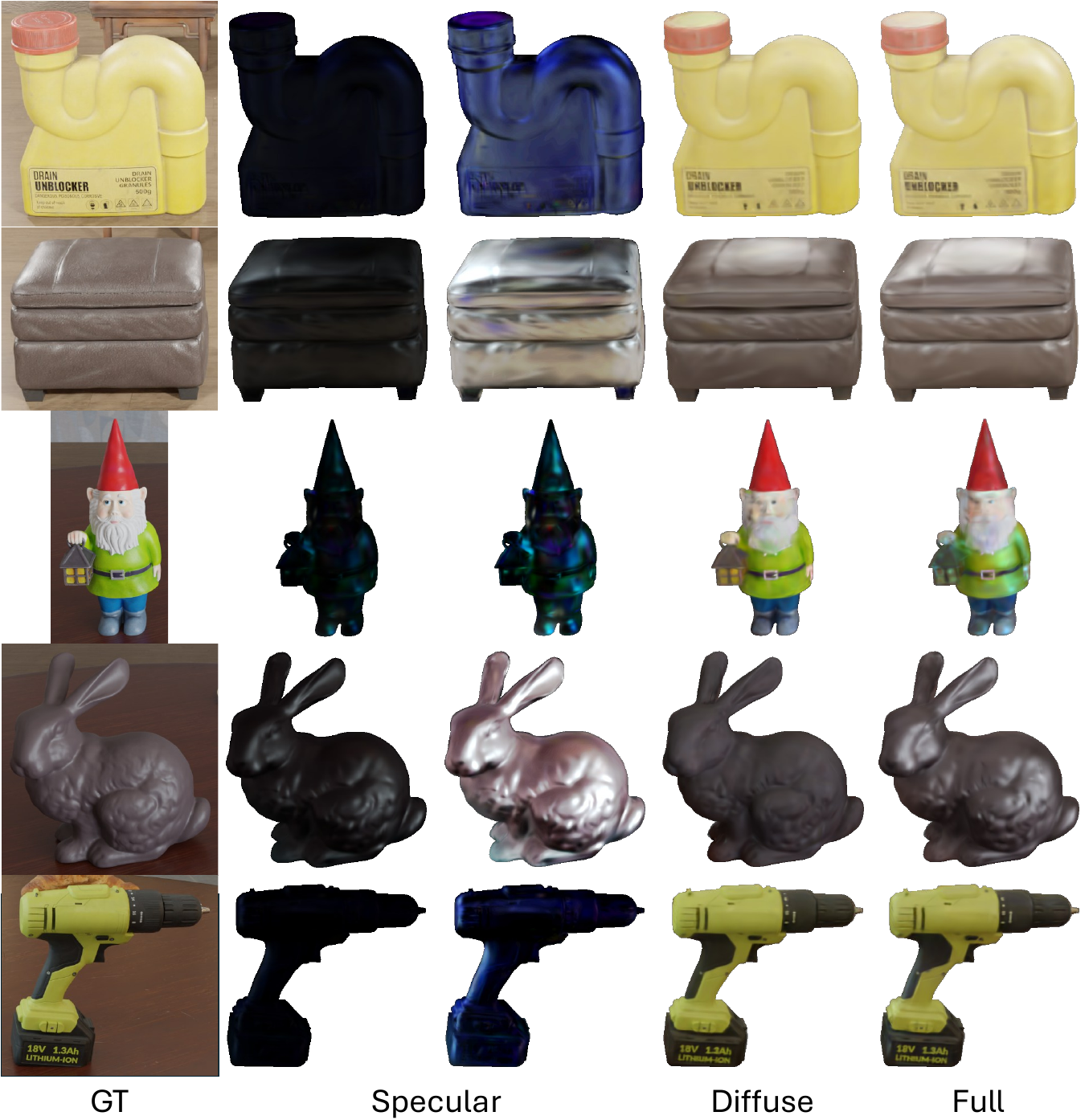}
  \caption{
    \textbf{Visualization of specular and diffuse components.} In the second column of the specular components, we show the results with increased brightness for better visibility. Our method successfully decomposes the object appearance into specular and diffuse components, except for the challenging case of the ottoman.
  }
  \label{fig:results_specular_diffuse_vis}
\end{figure*}

\begin{table*}[t]
\caption{
\textbf{Comparison with per-Gaussian SH, a variant of our method that optimizes SH coefficients independently for each Gaussian.}
We report (a) surface normal accuracy and (b) novel view synthesis quality on synthetic data.
Optimizing per-Gaussian SH can be unstable on complex objects (\eg, the bunny), leading to degraded performance.
}
\label{tab:ablation_per_gaussian_sh}
\centering
\small

\begin{subtable}[t]{\textwidth}
\caption{\textbf{Surface normal accuracy.} Mean/median/80th percentile angular error ($\downarrow$).}
\label{tab:ablation_motion_aware_appearance_normals}
\centering
\setlength{\tabcolsep}{2.pt}
\resizebox{\textwidth}{!}{%
\begin{tabular}{lrrrrrrrrrrrrrrrrrrrrr}
\toprule
\multirow{2}{*}{Method}
& \multicolumn{3}{c}{drain cleaner}
& \multicolumn{3}{c}{ottoman}
& \multicolumn{3}{c}{bunny}
& \multicolumn{3}{c}{garden gnome}
& \multicolumn{3}{c}{drill}
& \multicolumn{3}{c}{Average} \\
\cmidrule(lr){2-4}\cmidrule(lr){5-7}\cmidrule(lr){8-10}\cmidrule(lr){11-13}\cmidrule(lr){14-16}\cmidrule(lr){17-19}
& Mean $\downarrow$ & Med. $\downarrow$ & P80 $\downarrow$
& Mean $\downarrow$ & Med. $\downarrow$ & P80 $\downarrow$
& Mean $\downarrow$ & Med. $\downarrow$ & P80 $\downarrow$
& Mean $\downarrow$ & Med. $\downarrow$ & P80 $\downarrow$
& Mean $\downarrow$ & Med. $\downarrow$ & P80 $\downarrow$
& Mean $\downarrow$ & Med. $\downarrow$ & P80 $\downarrow$ \\
\midrule
per-Gaussian SH
& \textbf{9.2} & \textbf{5.8} & 13.8
& \textbf{17.3} & 12.6 & 19.1
& 12.8 & 10.2 & 18.1
& \textbf{18.5} & \textbf{13.7} & \textbf{29.0}
& 15.9 & 10.1 & 23.8
& \textbf{14.7} & 10.5 & 20.8 \\
Ours
& 9.9 & 7.3 & \textbf{13.6}
& 17.5 & \textbf{12.2} & \textbf{18.5}
& \textbf{12.2} & \textbf{9.2} & \textbf{16.8}
& 19.4 & 14.4 & 30.0
& \textbf{15.1} & \textbf{9.5} & \textbf{21.8}
& 14.8 & 10.5 & \textbf{18.5} \\
\bottomrule
\end{tabular}
}
\end{subtable}

\vspace{2mm}

\begin{subtable}[t]{\textwidth}
\caption{\textbf{Novel view synthesis accuracy.} PSNR/SSIM ($\uparrow$), L1/LPIPS ($\downarrow$).}
\label{tab:ablation_motion_aware_appearance_nvs}
\centering
\setlength{\tabcolsep}{3.0pt}

\resizebox{\textwidth}{!}{%
\begin{tabular}{lcccccccccccc}
\toprule
\multirow{2}{*}{Method}
& \multicolumn{4}{c}{drain cleaner}
& \multicolumn{4}{c}{ottoman}
& \multicolumn{4}{c}{bunny} \\
\cmidrule(lr){2-5}\cmidrule(lr){6-9}\cmidrule(lr){10-13}
& PSNR $\uparrow$ & L1 $\downarrow$ & SSIM $\uparrow$ & LPIPS $\downarrow$
& PSNR $\uparrow$ & L1 $\downarrow$ & SSIM $\uparrow$ & LPIPS $\downarrow$
& PSNR $\uparrow$ & L1 $\downarrow$ & SSIM $\uparrow$ & LPIPS $\downarrow$ \\
\midrule
per-Gaussian SH
& \textbf{41.75} & \textbf{0.00089} & \textbf{0.9962} & \textbf{0.0062}
& 30.93 & 0.00334 & 0.9900 & 0.0147
& 41.88 & 0.00110 & 0.9967 & 0.0063 \\
Ours
& 41.62 & 0.00098 & 0.9961 & 0.0065
& \textbf{32.77} & \textbf{0.00283} & \textbf{0.9904} & \textbf{0.0145}
& \textbf{43.99} & \textbf{0.00091} & \textbf{0.9969} & 0.0063 \\
\bottomrule
\end{tabular}
}

\vspace{2mm}

\resizebox{\textwidth}{!}{%
\begin{tabular}{lcccccccccccc}
\toprule
\multirow{2}{*}{Method}
& \multicolumn{4}{c}{garden gnome}
& \multicolumn{4}{c}{drill}
& \multicolumn{4}{c}{Average} \\
\cmidrule(lr){2-5}\cmidrule(lr){6-9}\cmidrule(lr){10-13}
& PSNR $\uparrow$ & L1 $\downarrow$ & SSIM $\uparrow$ & LPIPS $\downarrow$
& PSNR $\uparrow$ & L1 $\downarrow$ & SSIM $\uparrow$ & LPIPS $\downarrow$
& PSNR $\uparrow$ & L1 $\downarrow$ & SSIM $\uparrow$ & LPIPS $\downarrow$ \\
\midrule
per-Gaussian SH
& \textbf{40.41} & \textbf{0.00098} & \textbf{0.9959} & \textbf{0.0044}
& \textbf{46.27} & \textbf{0.00050} & \textbf{0.9976} & \textbf{0.0040}
& 40.25 & 0.00136 & \textbf{0.9953} & \textbf{0.0071} \\
Ours
& 39.86 & 0.00100 & 0.9953 & 0.0054
& 46.12 & 0.00051 & 0.9973 & 0.0043
& \textbf{40.87} & \textbf{0.00125} & 0.9952 & 0.0074 \\
\bottomrule
\end{tabular}
}
\end{subtable}

\end{table*}

\begin{figure*}[t]
  \centering
  \includegraphics[width=\linewidth]{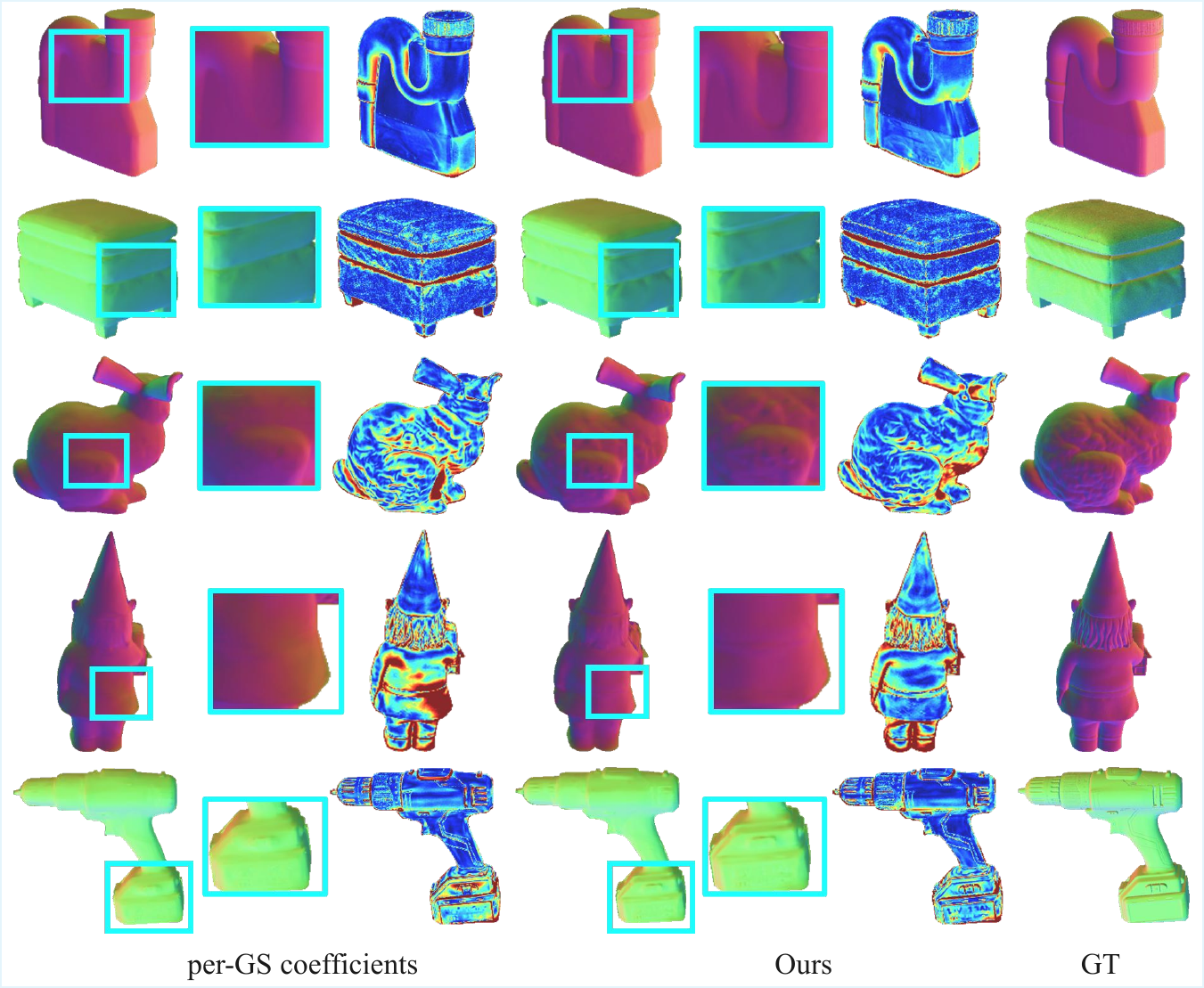}
  \caption{
    \textbf{Qualitative comparison with per-Gaussian SH.} In particular, our full model yields sharper reconstructions of fine surface details (\eg, the wrinkles on the bunny).
  }
  \label{fig:results_ablation_per_gaussian_sh}
\end{figure*}

\begin{figure*}[t]
  \centering
  \includegraphics[width=\linewidth]{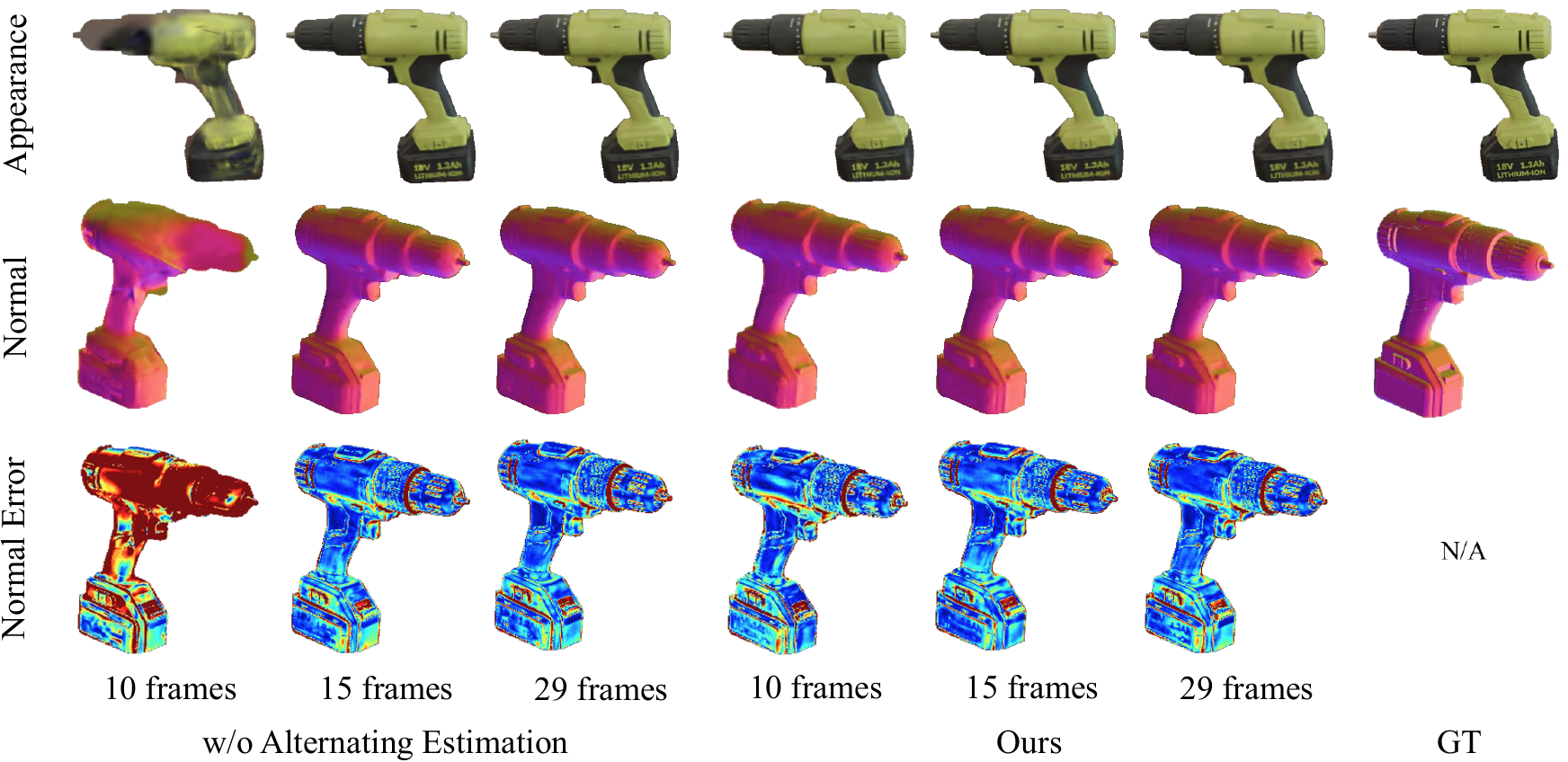}
  \caption{
    \textbf{Comparison with our method without alternating estimation (“w/o Alternating Estimation”).} In the “10 frames,” “15 frames,” and “29 frames” settings, the object rotates by approximately 39, 26, and 13 degrees, respectively, between adjacent frames. Our full method successfully recovers geometry and appearance even in the temporally sparsest setting.
  }
  \label{fig:results_ablation_alternating_estimation}
\end{figure*}

\begin{figure*}[t]
  \centering
  \includegraphics[width=\linewidth]{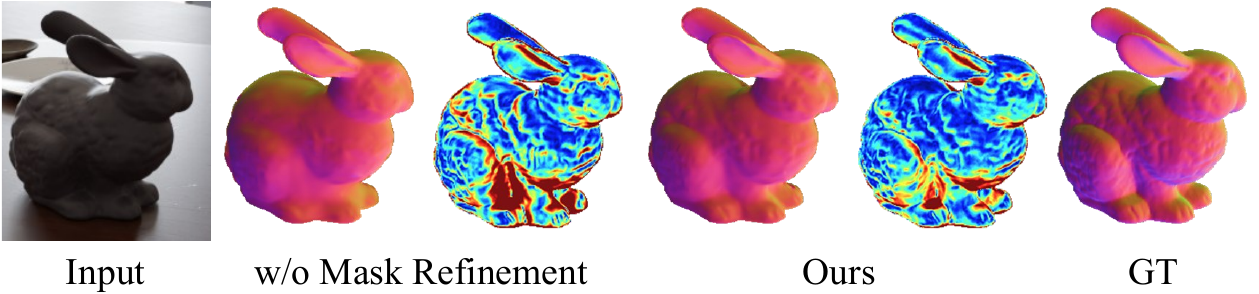}
  \caption{
    \textbf{Effectiveness of segmentation mask refinement.} We compare our full method with a variant without mask refinement (“w/o Mask Refinement”), where the per-Gaussian mask values are fixed during optimization. Refining the mask values jointly with geometry and appearance is essential for converging to the true fine-scale geometry.
  }
  \label{fig:results_ablation_mask_refinement}
\end{figure*}

\begin{figure*}[t]
  \centering
  \includegraphics[width=\linewidth]{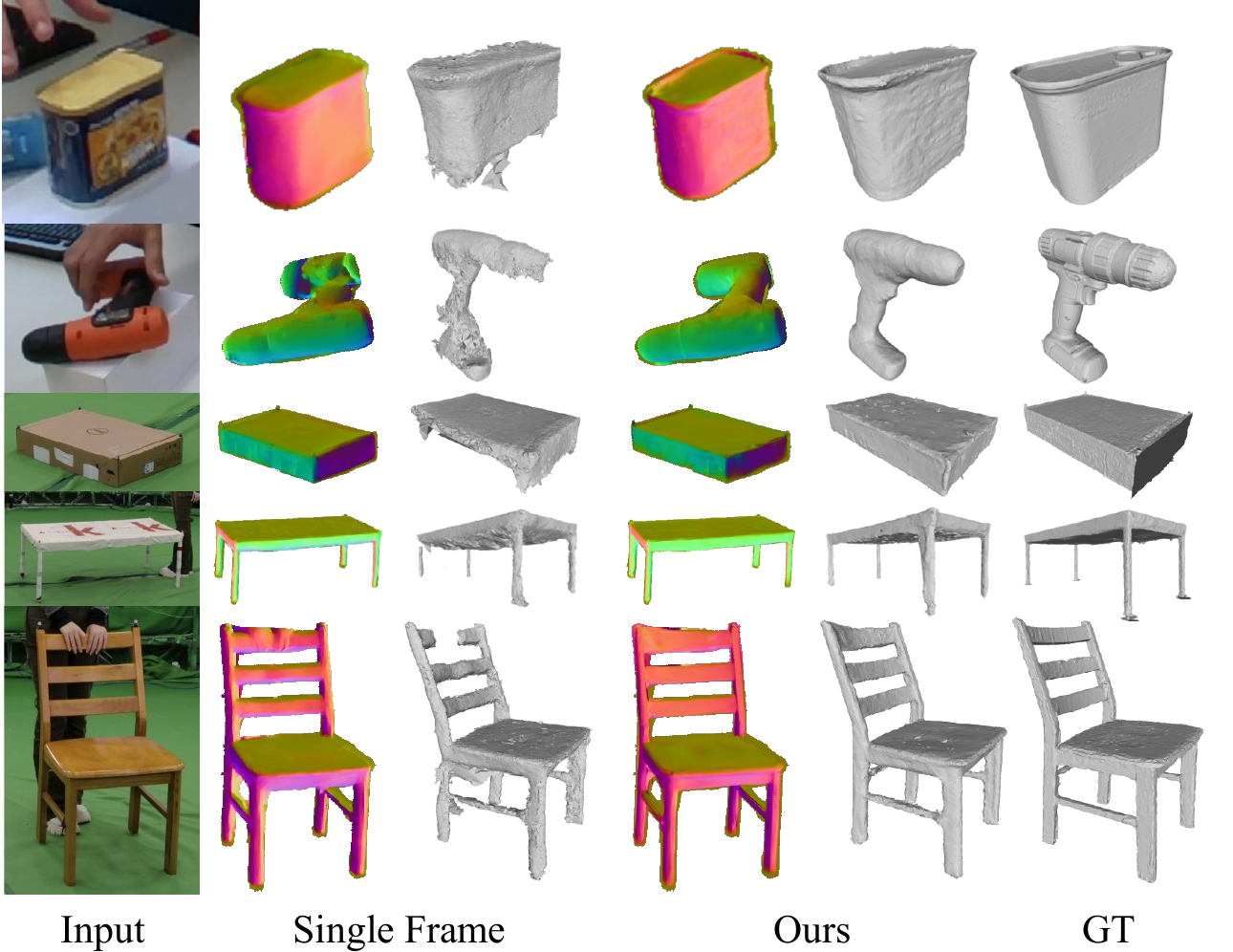}
  \caption{
    \textbf{Surface normals and mesh models recovered from real-world datasets.} The results demonstrate the effectiveness and robustness of our method across diverse human-object interaction scenarios.
  }
  \label{fig:results_geometry_real_supp}
\end{figure*}

\Cref{fig:results_normal_supp} shows additional qualitative results of the surface normals recovered from the synthetic dataset. The drain cleaner has relatively simple geometry and can be handled well by the baseline methods, leaving limited room for improvement by our method. 
Nevertheless, our method achieves surface-normal estimates comparable to the baselines even on such an object, indicating robustness across different object complexities.

\Cref{fig:results_appearance_supp} shows additional novel view synthesis results on the synthetic dataset. 
Our appearance model improves the fidelity of novel view synthesis across different objects. 
The ottoman (the second row) remains challenging even for our method, as it exhibits limited surface-normal variation, providing only sparse view-dependent appearance observations for appearance recovery.

\Cref{fig:results_comparison_dgmesh} shows a qualitative comparison with DG-Mesh~\cite{liu2024dynamic}. While it supports deformable objects, it struggles to recover fine surface details from sparse inputs captured by static cameras.

\subsection{Effectiveness of Motion-Aware Appearance Modeling}

\Cref{fig:results_ablation_diffuse_specular_normal,fig:results_ablation_diffuse_specular_appearance} show qualitative results of the ablation studies on the diffuse and specular components of our motion-aware appearance model.
Removing either component degrades the recovery of \ryosuke{view- and motion-dependent} effects, leading to less faithful novel view synthesis and reduced surface-normal quality on objects with challenging materials such as the bunny.

\Cref{fig:results_specular_diffuse_vis} shows visualizations of the estimated specular and diffuse components.
These results indicate that our method can effectively decompose the object appearance into specular highlights and diffuse shading, enabling separate analysis of each component.

We further validate our motion-aware appearance model by comparing it against \textit{per-Gaussian SH}, a variant of our method that optimizes spherical harmonics (SH) coefficients independently for each Gaussian, similar to 3D Gaussian Splatting (3DGS).
\Cref{tab:ablation_per_gaussian_sh,fig:results_ablation_per_gaussian_sh} show quantitative and qualitative results.
Optimizing per-Gaussian SH coefficients can be unstable on complex objects such as the bunny, leading to degraded surface-normal and novel-view synthesis accuracy.
In contrast, our full method with shared SH coefficients successfully recovers high-frequency details of surface normals, demonstrating the advantage of the shared SH coefficients.

\subsection{Effectiveness of Alternating Estimation}

To further evaluate the effectiveness of our alternating estimation framework, we test our method under more challenging temporally sparse settings. Specifically, we construct subsampled videos by skipping every other frame and every two frames. In these settings, the object rotates by approximately 26 degrees and 39 degrees, respectively, between adjacent frames.
\Cref{fig:results_ablation_alternating_estimation} shows qualitative comparisons between our full method and a variant without alternating estimation (“w/o Alternating Estimation”). Without alternating estimation, the joint estimation suffers from the strong interdependency between geometry, pose, and appearance. In contrast, our method successfully recovers geometry and appearance even from the sparsest videos, clearly demonstrating the effectiveness of the alternating estimation framework.

\subsection{Effectiveness of Segmentation Mask Refinement}
We also evaluate the effectiveness of refining the per-Gaussian segmentation mask values $m_i$.
We compare our method with \textit{w/o Segmentation Mask Refinement}, a variant in which the mask values are fixed to their initial estimates throughout optimization.
\Cref{fig:results_ablation_mask_refinement} shows qualitative results.
Mask refinement is essential for converging to the true fine-scale geometry.

\subsection{Additional Qualitative Results on Real-World Datasets}

\Cref{fig:results_geometry_real_supp} shows additional qualitative results of the reconstructed surface geometry on real-world datasets.
Our method recovers fine-grained geometric details across objects with diverse shapes and material properties.

\end{document}